\documentclass[10pt,twocolumn,letterpaper]{article}

\usepackage{wacv}              
\usepackage[accsupp]{axessibility}  

\definecolor{onlinegreen}{RGB}{0,120,110}

\newcommand{\cmark}{\ding{51}}

\usepackage{makecell}

\usepackage{pifont}
\usepackage[table]{xcolor}
\usepackage{booktabs,makecell}

\usepackage{graphicx}   

\usepackage{tabularx}

\definecolor{offgray}{HTML}{EFEFEF}   
\definecolor{onlblue}{HTML}{EDF4FF}   

\newcommand{\swatch}[1]{{\setlength{\fboxsep}{0pt}%
  \fcolorbox{black!25}{#1}{\rule{0pt}{1.5ex}\rule{3.0ex}{0pt}}}}

\definecolor{wacvblue}{rgb}{0.21,0.49,0.74}
\usepackage[pagebackref,breaklinks,colorlinks,allcolors=wacvblue]{hyperref}

\definecolor{DarkGreen}{RGB}{0,100,0}
\definecolor{DarkRed}{RGB}{139,0,0}
\usepackage{multirow}
\usepackage{graphicx}
\usepackage{subcaption}
\usepackage[most]{tcolorbox}
\usepackage{algorithm}
\usepackage{algpseudocode}

\usepackage[table]{xcolor}
\usepackage{booktabs}
\usepackage{multirow}
\usepackage{graphicx}
\usepackage{xspace}

\usepackage{menukeys}

\definecolor{secblue}{RGB}{232,232,247}
\definecolor{secgreen}{RGB}{231,245,232}
\definecolor{secyellow}{RGB}{247,247,225}
\definecolor{secred}{RGB}{250,232,232}
\definecolor{secgray}{RGB}{240,240,240}
\definecolor{lightgraytext}{RGB}{130,130,130}
\definecolor{darkgreen}{RGB}{0,100,0}

\usepackage{booktabs}
\usepackage{multirow}
\usepackage[table]{xcolor}
\usepackage{adjustbox}
\usepackage{pifont}
\newcommand{\xmark}{\ding{55}}

\definecolor{oursblue}{HTML}{EDF4FF}
\definecolor{groupgray}{HTML}{F7F7F7}
\definecolor{backbonegray}{HTML}{666666}

\newcommand{\eli}[1]{{\color{red} {[Elisa] #1}}}
\newcommand{\longc}[1]{\ifvmode\noindent\fi{\color{blue}#1}}

\newcommand{\methodFull}{\textbf{H}ierarchical \textbf{S}tep \textbf{D}ecomposition }
\newcommand{\methodShort}{HSD}

\def\wacvPaperID{*****} 
\def\confName{WACV}
\def\confYear{2027}

\title{Hierarchical Step Decomposition 
for Training-free Online Video Step Grounding}

\author{Long Zhou$^{1}$ \quad Luca Zanella$^{1}$ \quad Paolo Rota$^{1}$\\
Massimiliano Mancini$^{1}$ \quad Elisa Ricci$^{1,2}$\\
{\normalsize $^{1}$University of Trento \quad $^{2}$Fondazione Bruno Kessler}\\
{\normalsize \url{https://zhoulong0.github.io/hsd/}}
}

\begin{document}
\maketitle
\begin{abstract}
Given a set of procedural steps, Video Step Grounding (VSG) aims to identify which steps occur in a video and localize when they are performed. Existing methods are either offline, requiring access to the full video before making predictions, or online, but relying on fixed-length video segments producing noisy, over-segmented outputs. In this work, we propose \methodFull (\methodShort), a training-free framework for online VSG that leverages the semantic understanding capabilities of Large Multimodal Models. \methodShort{} addresses two central challenges in online grounding: constructing semantically coherent temporal segments from a streaming video, and bridging the semantic gap between coarse procedural step descriptions and visual observations. To construct elementary temporal units, \methodShort{} incrementally groups incoming segments into semantically consistent events, using an anchor caption that is continuously validated against the visual stream. Based on these events, we perform zero-shot procedural step classification and selectively refine uncertain predictions by decomposing candidate steps into fine-grained substeps that capture more nuanced visual evidence. By operating on semantically coherent events rather than every fixed-length segment, \methodShort{} reduces redundant computation and produces more stable predictions. Experiments on HT-Step, Ego4D Goal-Step, and CrossTask show that \methodShort{} substantially outperforms state-of-the-art methods, overall achieving better step recognition, more accurate temporal localization, and faster inference. 
\end{abstract}
    
\section{Introduction}
\begin{figure*}[ht]
    \centering
    \includegraphics[width=0.95\linewidth]{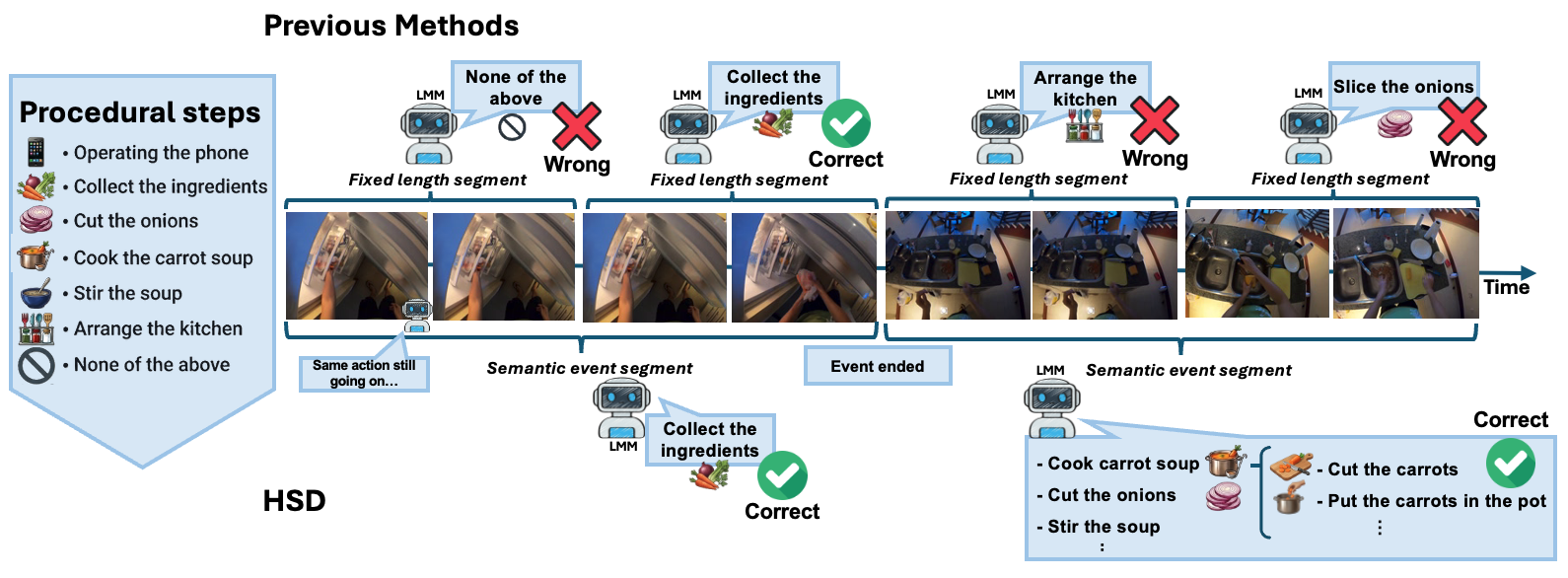}
    \caption{
   \textit{Top:} previous works for Online VSG~\cite{zanella2025baglm} predict procedural steps on fixed-length segments. \textit{Bottom:} \methodFull (\methodShort) constructs semantically coherent temporal event segments and, if necessary, decomposes procedural steps into multiple substeps to mitigate the granularity mismatch between coarse procedural step descriptions and visual observations.
    }
    \label{fig:teaser}
    \vspace{-4pt}
\end{figure*}

When humans perform a task (e.g., cooking), they undergo a series of steps (e.g., cutting, adding ingredients, mixing): recognizing these steps is essential for applications that analyze, monitor, or assist humans. 
To this end, given a video and a list of procedural steps in natural language, Video Step Grounding (VSG) \cite{Zhukov2019, mavroudi2023vina}, aims to 
identify which steps are present in the video and when they occur. 


Recent progress in VSG has been largely driven by weakly supervised approaches~\cite{han2022tan,mavroudi2023vina,chen2024mptva,li2024nasva} that formulate temporal grounding as alignment between textual descriptions and predefined video units, such as individual frames or fixed-length segments. While effective, these methods rely on task-specific supervision on curated training data, and typically process the video offline. As a result, they are difficult to transfer to unseen tasks and not well suited for streaming scenarios, such as always-on assistive devices, where a video must be processed incrementally as it is observed. This limitation has recently motivated the development of \emph{online}, \emph{training-free} approaches based on Large Multimodal Models (LMMs)~\cite{zanella2025baglm}.

Online Video Step Grounding (OVSG) is significantly more difficult, as it requires recognizing a step given only past frames.
A major challenge lies in the choice of the temporal units to use for prediction. The only existing OVSG method~\cite{zanella2025baglm} relies on fixed-length segments, a rigid representation poorly aligned with instructional procedures, where action lengths vary substantially.
For instance, short segments often lack sufficient visual evidence to reliably identify a step while longer segments may encompass multiple actions, mixing visual evidence from different steps and leading to ambiguous predictions. 
A second challenge arises from the semantic gap between procedural step descriptions and local visual observations. Step descriptions are often coarse and goal-oriented. For instance, a step such as \textit{cook carrot soup} may involve several finer-grained actions, such as \emph{taking the carrots}, \emph{cutting them}, and \emph{placing them into a pot}. At any given timestep, the video may show only one of these sub-actions, so the visual information may only partially match the step description. As a result, the model may miss the correspondence with the higher-level step or may confuse it with visually similar actions.

In this work, we propose \methodFull (\methodShort), a training-free framework for OVSG that jointly addresses both challenges. \methodShort{} operates in three stages. In the first, instead of grounding predictions on fixed-length temporal units as in~\cite{zanella2025baglm}, we incrementally build semantically consistent event segments by determining whether each incoming visual observation continues the current event or initiates a new one. 
As a result, the model performs grounding over semantically coherent video units, avoiding the rigidity and the noise of fixed-length partitioning. 
In the second stage, we perform an initial prediction of the procedural step associated with an event, leveraging the zero-shot capabilities of an LMM~\cite{bai2025qwen25vltechnicalreport,bai2025qwen3vltechnicalreport,chen2025internvl25, wang2025internvl3}. To reduce the granularity mismatch between the coarse procedural steps and local visual evidence, we introduce a third stage, \textit{i.e.}, a hierarchical step refinement stage, that is activated when the model has low confidence in the initial event-to-step assignment. In this stage, coarse descriptions of procedural steps are expanded into multiple, fine-grained substeps, and the current event is re-evaluated against these more detailed alternatives. {Importantly, all stages of \methodShort{} use the same, frozen LMM, in a unified training-free pipeline.} 
A comparison between \methodShort{} and previous methods is illustrated in Fig. \ref{fig:teaser}.

We evaluate \methodShort{} on HT-Step \cite{afouras2023htstep}, Ego4D Goal-Step \cite{song2023ego4dgs} and CrossTask \cite{Zhukov2019}.
Overall, our approach advances the state of the art in OVSG by jointly improving step recognition, temporal localization, and inference efficiency. Using the same Qwen2.5-VL-7B backbone, it runs approximately $3\times$ faster on short videos and up to $16\times$ faster on longer videos than the prior state-of-the-art method, while improving average mAP by $+6.98$ on HT-Step and $+6.63$ on CrossTask, and increasing R@1 at mIoU$=0.3$ by $+6.42$ points on Ego4D Goal-Step.

\textbf{Contributions.} We make three key contributions: 1) We introduce a training-free framework for online VSG that incrementally constructs semantically consistent event segments. 2) We propose a hierarchical refinement mechanism that resolves granularity mismatches between textual descriptions of coarse procedural steps and local visual observations. 3) We evaluate \methodShort{} on instructional video benchmarks, showing that it significantly outperforms state-of-the-art methods in recognition, localization, and efficiency. 
\section{Related Work}

Early work on VSG considered weak supervision from narrations and task structure. CrossTask~\citep{Zhukov2019} exploited ordered steps and narrations to learn shared step classifiers across tasks, while TAN~\citep{han2022tan} and VINA~\citep{mavroudi2023vina} learned narration-to-video alignment from pseudo-labels and wikiHow articles respectively. MPTVA~\citep{chen2024mptva} and NaSVA~\citep{li2024nasva} improved the alignment by refining narrations with a Large Language Model (LLM), through filtered summaries and by converting ASR narrations into descriptive steps. More recently, BAGLM~\citep{zanella2025baglm} introduced a training-free approach for OVSG, leveraging LMMs over fixed-duration video segments and refining predictions with Bayesian filtering. While \methodShort{} is also online and training-free, it fundamentally differs from BAGLM as it groups video segments and incorporates a hierarchical step-to-substep refinement, enabling more accurate predictions on fine-grained actions.

VSG is also related to several other video understanding tasks.
Temporal action localization (TAL) and segmentation (TAS)
detect action boundaries and assign frame-level labels from a fixed vocabulary
shared between training and test time, 
offline~\cite{farha2019ms,shi2023tridet,yi2021asformer,zhang2022actionformer}
or, more recently,
online~\cite{kang2021cagil,kim2022sliding,reza2024hat,song2024matr,shen2024protas,zhong2024ontas}.
Their closed-set formulation, however, requires retraining if new actions are introduced. Open-vocabulary TAL~\cite{liberatori2024t3al,han2025freezad,nag2022stale}
lifts this constraint and localizes actions unseen during training, but 
typically operates offline, specifies the action of interest through category
labels rather than free-form text, and requires a separate inference pass per
action. Dense video captioning~\cite{yang2023vid2seq,yang2023vidchapters,guo2025trace}
and its streaming variants~\cite{zhou2024streamingdvc} produce free-form captions
with associated temporal boundaries, but are not conditioned on a given list of
procedural steps. Video temporal
grounding~\cite{zhang2021vslnet,lei2021detecting,lin2023univtg}, including 
temporal-grounding video LLMs~\cite{ren2024timechat,pramanick2025enrich,li2026universal} and causal variants~\cite{gan2023tsgsv,zheng2025hierarchical},
returns a temporal interval for a textual query, yet assumes the query occurs in the video offering no mechanism to abstain when the queried action is
absent. Generic event boundary detection~\cite{shou2021generic,jung2025ogebd,mounir2023streamer} produces taxonomy-free boundaries, close in spirit to our segments grouping, but learns them from boundary annotations rather than training-free. OVSG differs from all of these in requiring online, open-vocabulary, single-pass predictions over free-form step descriptions, without assuming every step occurs in the video.
\section{Method}
\label{sec:method}

\begin{figure*}[t]
    \centering
    \includegraphics[width=0.95\linewidth]{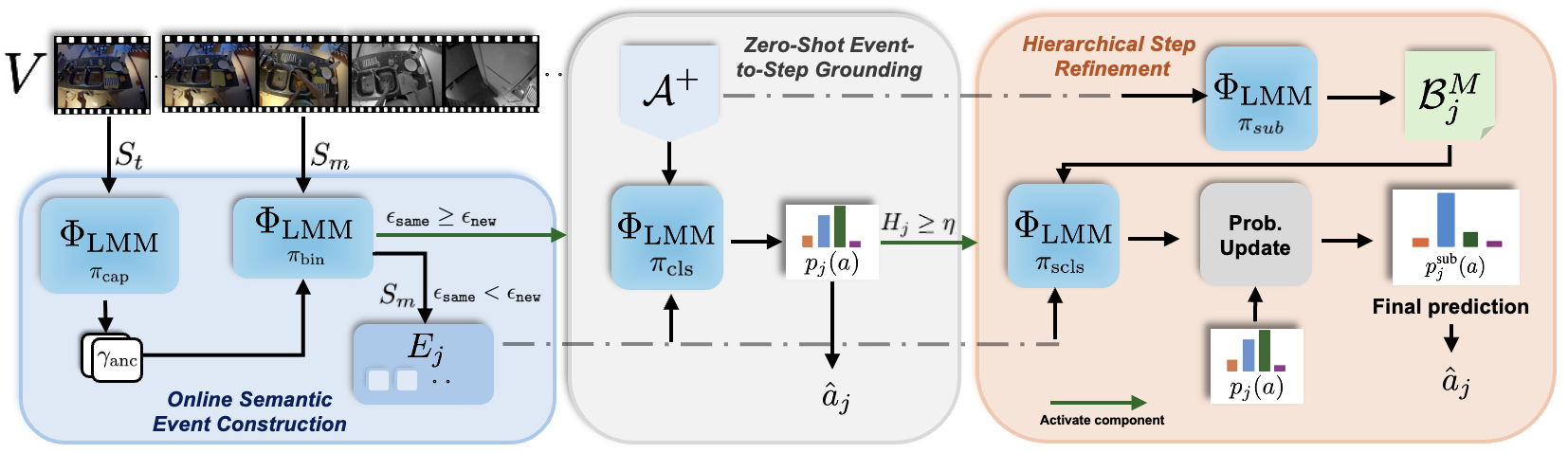}
    \caption{\textbf{Overview of \methodShort{}.} 
    Given an input video $V$, the Online Semantic Event Construction module incrementally builds each event $E_j$ from fixed-duration segments using a frozen LMM $\Phi_{\mathrm{LMM}}$. The first segment of $E_j$ is captioned as $\gamma_{\mathrm{anc}}$, and each arriving segment is appended to $E_j$ if its continuity score $\epsilon_{\texttt{same}}$ exceeds $\epsilon_{\texttt{new}}$ ; otherwise, $E_j$ is terminated and a new event is initialized. For each completed event $E_j$, the Zero-Shot Event-to-Step Grounding module predicts a probability distribution $p_j$ over procedural steps. If its entropy $H_j$ falls below threshold $\eta$, the most likely step is predicted. Otherwise, Hierarchical Step Refinement decomposes the top-$k$ steps into substeps, re-evaluates $E_j$, aggregates substep scores into $p^{\mathrm{sub}}_j$, and outputs the final prediction $\hat{a}_j$.}
    \label{fig:htstep}
\end{figure*}

In this section, we illustrate 
the proposed training-free approach for
OVSG. We first formally introduce the OVSG problem (Sec. \ref{sec:problem-formulation}). We then describe the first building block of our approach: a simple yet effective technique for identifying continuous segments corresponding to the same semantic event (Sec. \ref{sec:event-construction}). Then, we detail how we assign semantic labels to each event via visual question answering (Sec.~\ref{sec:event-grounding}), leveraging sub-step decomposition to refine uncertain predictions (Sec.~\ref{sec:hierarchical-refinement}).

\subsection{Online Video Step Grounding}
\label{sec:problem-formulation}
Given a video depicting a sequence of procedural {steps}, Video Step Grounding  aims to identify which steps are performed and localize their temporal boundaries within the video. Formally, let $V \in \mathcal{V}$ denote a video in the visual space $\mathcal{V}$, and $\mathcal{A} = \{a_i\}_{i=1}^{M}$ denote a given set of $M$ candidate procedural steps, where each $a_i \in \mathcal{L}$ is a natural language description in the space $\mathcal{L}$.  We model the video as a sequence of $T$ non-overlapping segments $\mathcal{S} = \{S_t\}_{t=1}^{T}$, where $S_t \subset V$ is the $t$-th incoming segment. In the online setting~\cite{zanella2025baglm}, predictions at time $t \leq T$ are restricted to the partial history $\mathcal{S}_{1:t} = \{S_1, \dots, S_t\}$, rather than the full sequence $\mathcal{S}$ available in offline inference. For each segment $\mathcal{S}_{t}$, the model outputs a prediction $\hat{a}_t \in \mathcal{A} \cup \{\varnothing\}$, where $\varnothing$ denotes that none of the candidate procedural steps is being performed. A step prediction should be made as soon as enough evidence is available, without access to future frames, as in online temporal action detection.~\citep{kang2021cagil,reza2024hat,song2024matr,kim2022sliding}.

Previous work~\cite{zanella2025baglm} addressed online VSG by estimating, for each incoming fixed-length segment $S_t$, a probability distribution over the candidate procedural steps in 
$\mathcal{A}^+=\mathcal{A} \cup \{\varnothing\}$. While effective, this formulation has two key limitations. 
First, operating on fixed temporal units leads to semantic noise, as action boundaries rarely align with uniform segments.
Second, it assumes that the descriptions associated with procedural steps are sufficient to 
distinguish among visually similar actions. This assumption often breaks if the visual content is ambiguous (\eg, \textit{holding a knife over a cutting board}), and multiple steps may remain plausible (\eg, \textit{slicing onions} or \textit{cutting vegetables}).

In the following, we describe how both challenges can be addressed by leveraging the zero-shot reasoning capabilities of LMMs to (i) construct semantic events; (ii) perform zero-shot predictions; and (iii) refine them in case of uncertainty (Fig.~\ref{fig:htstep}). 
Given the increasing granularity of the approach (\textit{i.e.}, task$\rightarrow$procedural steps$\rightarrow$substeps), we name our method \methodFull (\textbf{\methodShort}). In the following we describe its three main stages.

\subsection{Online Semantic Event Construction}
\label{sec:event-construction}

The first stage of \methodShort{} converts a stream of fixed-duration input segments into adaptive temporal \textit{events}. We denote as \textit{event}  a  
sequence of segments with consistent semantic. 
In this stage, as in subsequent ones, we employ a Large Multimodal Model $\Phi_{\mathrm{LMM}}:\mathcal{V}\times\mathcal{L}\rightarrow\mathcal{L}$, which maps a visual input and a textual prompt to a language output. The model is used in a frozen, training-free setting. 

When the video or a new event begins, 
the incoming segment $S_t$ is selected as the \emph{anchor segment}. We perform two operations. First, the LMM generates a caption $\gamma_{\mathrm{anc}} = \Phi_{\mathrm{LMM}}(S_t, \pi_{\mathrm{cap}})$, where $\pi_{\mathrm{cap}}$ is a captioning prompt\footnote{All prompts are provided in Appendix~\ref{app:prompts}}. This caption provides a compact textual description of the current event and it is used as an \textit{anchor} to identify subsequent segments associated with the same action. 
Second, we initialize the active event as $E_j=\{S_t\}$, where $E_j$ denotes the ordered set of fixed-length segments assigned to the $j$-th event.

For each subsequent incoming segment $S_m$, with $m > t$, we assess its continuity with the active event by comparing it against the anchor caption. Specifically, we prompt the LMM with a binary query $\pi_{\mathrm{bin}}(\gamma_{\mathrm{anc}})$, which asks whether the visual content of $S_m$ corresponds to the \textit{same} event described by $\gamma_{\mathrm{anc}}$ or marks the start of a \textit{new} one.
We define the continuity score $\epsilon_{\texttt{same}}$ as the probability that the model assigns to the \texttt{same} outcome under its next-token distribution:
$\epsilon_{\texttt{same}} =
P_{\Phi_{\mathrm{LMM}}}\!\left(\texttt{same} \mid S_m, \pi_{\mathrm{bin}}(\gamma_{\mathrm{anc}})\right)$.  Similarly, we denote as $\epsilon_{\texttt{new}}=P_{\Phi_{\mathrm{LMM}}}\!\left(\texttt{new} \mid S_m, \pi_{\mathrm{bin}}(\gamma_{\mathrm{anc}})\right)$ the probability assigned by $\Phi_{\mathrm{LMM}}$ to \texttt{new}. 
The segment is considered part of the active event if $\epsilon_{\texttt{same}}\geq \epsilon_{\texttt{new}}$. 
In this case, the event is extended through temporal concatenation: $E_j \leftarrow E_j \cup \{S_m\}$. Otherwise, the current event is considered complete and passed to the next stage (Sec.~\ref{sec:event-grounding}). 
The incoming segment $S_m$ then initializes a new event $E_{j+1}$, and the process repeats. This recursive online procedure partitions the video stream into semantically coherent events whose duration adapts to the underlying actions, providing more coherent inputs to the grounding stage. We report the pseudo-code in Appendix~\ref{app:method_details}.

\subsection{Zero-Shot Event-to-Step Grounding}
\label{sec:event-grounding}

After constructing a semantic event $E_j$, we leverage the zero-shot recognition capabilities of $\Phi_{\mathrm{LMM}}$ to identify which procedural step it depicts. Because some events may capture background activities or out-of-procedure actions, we use the extended set of candidates 
$\mathcal{A}^{+} = \mathcal{A} \cup \{\varnothing\}$. 

We frame event-to-step grounding as a multiple-choice question answering task. Let $\pi_{\mathrm{cls}}(\mathcal{A}^{+})$ denote a classification prompt that provides the overall task context (\textit{e.g.}, ``\textit{Making a sandwich}'') and the enumerated list of candidates in $\mathcal{A}^{+}$. For each candidate $a \in \mathcal{A}^{+}$, we use the probability assigned by $\Phi_{\mathrm{LMM}}$ to its corresponding answer token, \textit{i.e.}, the next-token probability of the multiple-choice letter associated with $a$:
\begin{equation}
\label{eq:zero-shot}
\hat{p}_j(a) = P_{\Phi_{\mathrm{LMM}}}\!\left(a \mid E_j, \pi_{\mathrm{cls}}(\mathcal{A}^{+})\right).
\end{equation}
We then normalize these scores over all candidates to obtain a probability distribution $p_j$ over $\mathcal{A}^{+}$:
\begin{equation}
\label{eq:normalization}
p_j(a) = \frac{\hat{p}_j(a)}{\sum_{a' \in \mathcal{A}^{+}} \hat{p}_j(a')}, \qquad a \in \mathcal{A}^{+}.
\end{equation}

The resulting distribution $p_j$ provides the initial semantic prediction for the completed event $E_j$. The class with the highest probability represents the predicted step, while the overall confidence determines whether the prediction requires further hierarchical refinement. 

\subsection{Hierarchical Step Refinement}
\label{sec:hierarchical-refinement}
While the initial distribution $p_j$ provides a semantic prediction for event $E_j$, the observed event may contain only partial visual evidence of the corresponding procedural step. This is particularly challenging when candidate steps in $\mathcal{A}$ are high-level and goal-oriented, while the event captures only a fine-grained sub-action, creating a mismatch in semantic granularity. To reduce this mismatch, we refine uncertain predictions by decomposing the most likely candidate steps into atomic, open-vocabulary, substeps (\textit{e.g.}, 
\textit{Make mashed potatoes} $\rightarrow$ \textit{boil the potatoes}, \textit{drain the potatoes}, \textit{mash the potatoes}) and re-evaluating the event segment at this finer level of granularity.

We first quantify the uncertainty of the initial prediction using the Shannon entropy of $p_j$:
\begin{equation}
H_j = -\sum_{a \in \mathcal{A}^{+}} p_j(a)\log p_j(a).
\end{equation}


For a predefined threshold $\eta$, if $H_j < \eta$ the prediction is considered confident, and we skip refinement and directly output
$\hat{a}_j=\arg\max_{a\in\mathcal{A}^{+}} p_j(a)$. However, if $H_j \geq \eta$, we need to further
disambiguate the prediction via hierarchical refinement. For computational
efficiency, this is restricted to the top-$k$ most plausible steps under the initial
distribution, $\mathcal{A}^{k}_j = \{a \in \mathcal{A}^{+} \mid p_j(a) \in
\operatorname{top}_k(p_j)\}$.

For each selected step $a \in \mathcal{A}^{k}_j$, we prompt $\Phi_{\mathrm{LMM}}$ using a substep generation prompt $\pi_{\mathrm{sub}}$ to decompose it into a set of finer-grained action descriptions:

\begin{equation}
\mathcal{B}_a = \Phi_{\mathrm{LMM}}(a, \pi_{\mathrm{sub}}) = \{b^a_{1},\dots,b^a_{L_a}\},
\end{equation}

where $L_a$ is the number of substeps identified for $a$. These substeps provide more atomic action descriptions that can be matched more directly to the visual evidence in $E_j$\footnote{Note that, for efficiency, we extract substeps only once, before processing the videos.}.  Although this operation only requires textual reasoning, we use $\Phi_{\mathrm{LMM}}$ instead of introducing an additional LLM, keeping the entire pipeline based on a single frozen LMM. 

We aggregate the generated substeps into a candidate pool $\mathcal{B}^{k}_j = \bigcup_{a\in \mathcal{A}^{k}_j} \mathcal{B}_a$ and perform zero-shot semantic prediction (as in Eq.~\eqref{eq:zero-shot}) over this finer-grained set. For each substep $b \in \mathcal{B}^{k}_j$, we compute its alignment score as
\begin{equation}
\hat{p}_j(b) = P_{\Phi_{\mathrm{LMM}}}\!\left(b \mid E_j, \pi_{\mathrm{scls}}(\mathcal{B}^{k}_j)\right),
\end{equation}
where $\pi_{\mathrm{scls}}$ is a multiple-choice classification prompt, now applied to the substep candidates. 
Next, we aggregate these substep scores back to the original candidate step space. Since a high-level step is supported if any of its atomic substeps are strongly detected, each candidate step $a \in \mathcal{A}^{+}$ inherits the maximum score of its constituent substeps:
\begin{equation}
\label{eq:substep_aggregation}
\hat{p}^{\mathrm{sub}}_j(a) = 
\begin{cases}
\max_{b\in \mathcal{B}_a} \hat{p}_j(b), & \text{if } a \in \mathcal{A}^{k}_j,\\
0, & \text{otherwise}.
\end{cases}
\end{equation}

Finally, we want fine-grained predictions to remain anchored to the broader procedural context, preventing the model from favoring a substep that is visually plausible but conceptually disconnected from the initial step prior. Thus, we weight the aggregated substep scores by their initial probability $p_j(a)$. The final refined probability distribution is obtained via normalization:
\begin{equation}
\label{eq:refined-normalization}
p^{\mathrm{sub}}_j(a) =
\frac{\hat{p}^{\mathrm{sub}}_j(a) \cdot p_j(a)}{\sum_{a' \in \mathcal{A}^{+}} \hat{p}^{\mathrm{sub}}_j(a') \cdot p_j(a')},
\qquad a \in \mathcal{A}^{+}.
\end{equation}

The final prediction in case of high entropy (\textit{i.e.}, $H_j \ge \eta$) is thus $\hat{a}_j=\arg\max_{a\in\mathcal{A}^{+}} p^\mathrm{sub}_j(a)$.

\begin{table*}[t]
\centering
\scriptsize
\setlength{\tabcolsep}{6.7pt}
\renewcommand{\arraystretch}{1.08}
\caption{
Comparison with state-of-the-art methods on HT-Step, CrossTask, and Ego4D Goal-Step.
Methods marked \swatch{onlblue} operate \emph{online} while methods marked \swatch{offgray} are \emph{offline}.}
\label{tab:main-sota}
\resizebox{\textwidth}{!}{%
\begin{tabular}{lcccc cccc c}
\toprule
\multirow{2}{*}{\textbf{Method}}
&
\multicolumn{4}{c}{\textbf{HT-Step mAP@IoU $\uparrow$}}
&
\multicolumn{4}{c}{\textbf{CrossTask mAP@IoU $\uparrow$}}
&
\multirow{2}{*}{%
  \raisebox{-3.0ex}{%
    \shortstack[c]{%
      \textbf{Ego4D Goal-Step}\\
      \textbf{R@1, IoU 0.3 $\uparrow$}
    }%
  }%
} \\
\addlinespace[-1.0\aboverulesep]
\cmidrule(lr){2-5}
\cmidrule(lr){6-9}
\addlinespace[-1.0\aboverulesep]
&
\textbf{@0.3} & \textbf{@0.5} & \textbf{@0.7} & \textbf{@[0.3--0.7]}
& \textbf{@0.3} & \textbf{@0.5} & \textbf{@0.7} & \textbf{@[0.3--0.7]}
& \\
\midrule
\multicolumn{10}{l}{\textit{Oracle}} \\
\addlinespace[1pt]
\rowcolor{offgray}
InternVL2.5-8B~\cite{chen2025internvl25}
& 34.15 & 26.09 & 21.87 & 27.02
& 45.40 & 36.33 & 28.92 & 36.65
& 30.89 \\
\rowcolor{offgray}
Qwen2.5-VL-7B~\cite{bai2025qwen25vltechnicalreport}
& 41.24 & 32.83 & 28.04 & 33.76
& 48.26 & 38.89 & 31.26 & 39.23
& 32.07 \\
\rowcolor{offgray}
Qwen3-VL-8B~\cite{bai2025qwen3vltechnicalreport}
& 38.91 & 30.61 & 26.12 & 31.52
& 50.73 & 42.01 & 34.81 & 42.35
& 48.59 \\
\addlinespace[1pt]
\midrule
\multicolumn{10}{l}{\textit{Dense video captioning (DVC)}} \\
\addlinespace[1pt]
\rowcolor{offgray}
Vid2Seq$_{\textcolor{lightgraytext}{\tiny \text{text}}}$~\cite{yang2023vid2seq}
& 9.89 & 2.46 & 0.43 & 3.86
& 12.16 & 2.83 & 0.24 & 4.56
& 0.48 \\
\rowcolor{offgray}
Vid2Seq$_{\textcolor{lightgraytext}{\tiny \text{boundary}}}$~\cite{yang2023vid2seq}
& 19.03 & 7.21 & 2.43 & 9.01
& 21.03 & 7.61 & 1.67 & 9.48
& 14.46 \\
\rowcolor{offgray}
VidChapters-7M$_{\textcolor{lightgraytext}{\tiny \text{text}}}$~\cite{yang2023vidchapters}
& 7.85 & 2.33 & 0.37 & 3.19
& 9.72 & 2.23 & 0.19 & 3.64
& 0.48 \\
\rowcolor{offgray}
VidChapters-7M$_{\textcolor{lightgraytext}{\tiny \text{boundary}}}$~\cite{yang2023vidchapters}
& 17.53 & 7.04 & 2.22 & 8.45
& 19.13 & 6.90 & 1.50 & 8.63
& 14.67 \\
\rowcolor{offgray}
TRACE$_{\textcolor{lightgraytext}{\tiny \text{text}}}$~\cite{guo2025trace}
& 9.50 & 3.12 & 0.90 & 4.20
& 6.77 & 1.54 & 0.23 & 2.53
& 1.19 \\
\rowcolor{offgray}
TRACE$_{\textcolor{lightgraytext}{\tiny \text{boundary}}}$~\cite{guo2025trace}
& 11.66 & 3.93 & 1.38 & 5.24
& 9.07 & 2.92 & 0.70 & 3.89
& 9.36 \\
\addlinespace[1pt]
\midrule
\multicolumn{10}{l}{\textit{Temporal sentence grounding}} \\
\addlinespace[1pt]
\rowcolor{offgray}
UniTime-7B~\cite{li2026universal}
& 19.92 & 11.48 & 5.52 & 12.27
& 28.86 & 18.90 & 8.17 & 18.68
& 24.87 \\
\midrule
\multicolumn{10}{l}{\textit{Video Step Grounding}} \\
\addlinespace[1pt]
\rowcolor{offgray}
TAN~\cite{han2022tan}
& 14.07 & 5.28 & 1.25 & 6.48
& 12.35 & 4.00 & 0.58 & 5.14
& 4.96 \\
\rowcolor{offgray}
NaSVA~\cite{li2024nasva}
& 15.02 & 5.70 & 1.11 & 6.89
& 17.42 & 5.69 & 0.86 & 7.32
& 10.41 \\
\addlinespace[1pt]
\rowcolor{onlblue}
NaSVA-Online~\cite{li2024nasva}
& 12.50 & 5.00 & 1.20 & 6.30
& 15.75 & 6.13 & 1.18 & 7.27
& 7.07 \\
\rowcolor{onlblue}
BAGLM$_{\textcolor{lightgraytext}{\tiny \text{InternVL2.5-8B}}}$~\cite{zanella2025baglm}
& 15.90 & 6.70 & 1.50 & 7.70
& 28.18 & 12.18 & 1.45 & 13.26
& 16.80 \\
\rowcolor{onlblue}
BAGLM$_{\textcolor{lightgraytext}{\tiny \text{Qwen2.5-VL-7B}}}$~\cite{zanella2025baglm}
& 15.80 & 6.60 & 1.70 & 7.70
& 24.58 & 10.22 & 1.47 & 11.43
& 13.09 \\
\rowcolor{onlblue}
BAGLM$_{\textcolor{lightgraytext}{\tiny \text{Qwen3-VL-8B}}}$~\cite{zanella2025baglm}
& 16.70 & 7.20 & 1.80 & 8.40
& 29.35 & 13.02 & 1.58 & 14.08
& 14.52 \\
\addlinespace[1pt]
\midrule
\rowcolor{onlblue}
\textbf{\methodShort}$_{\textcolor{lightgraytext}{\tiny \text{InternVL2.5-8B}}}$
& 23.98 & 11.60 & 4.73 & 13.15
& 33.41 & 17.65 & 6.10 & 18.80
& 20.33 \\
\rowcolor{onlblue}
\textbf{\methodShort}$_{\textcolor{lightgraytext}{\tiny \text{Qwen2.5-VL-7B}}}$
& 25.92 & 13.43 & 5.74 & 14.68
& 32.44 & 16.78 & 5.71 & 18.06
& 19.51 \\
\rowcolor{onlblue}
\textbf{\methodShort}$_{\textcolor{lightgraytext}{\tiny \text{Qwen3-VL-8B}}}$
& \textbf{26.45} & \textbf{13.78} & \textbf{6.16} & \textbf{15.13}
& \textbf{38.64} & \textbf{21.42} & \textbf{7.66} & \textbf{22.43}
& \textbf{25.36} \\
\bottomrule
\end{tabular}%
}
\end{table*}

\section{Experiments}
In this section, we present our evaluation protocol and implementation details (Sec.~\ref{sec:expsetup}). We then compare \methodShort{} with state-of-the-art methods in Sec.~\ref{sec:sota}, followed by ablation studies in Sec.~\ref{sec:ablation}.

\subsection{Evaluation protocol}
\label{sec:expsetup}
\noindent \textbf{Benchmarks.}
We evaluate \methodShort{} on three public benchmarks: HT-Step~\cite{afouras2023htstep}, Ego4D Goal-Step~\cite{song2023ego4dgs} and CrossTask~\cite{Zhukov2019}. HT-Step defines an evaluation setting with both seen and unseen activities. Following prior work~\cite{zanella2025baglm, mavroudi2023vina}, we use the seen validation split, since the official evaluation server for the seen and unseen test sets is currently unavailable. This split contains 600 videos spanning 120 activities, with 5 videos per activity. Ego4D Goal-Step, a subset of Ego4D~\cite{Grauman2022ego4d}, contains 851 egocentric videos with an average duration of about 26 minutes. Following \cite{zanella2025baglm, song2023ego4dgs}, we consider the validation set containing 134 videos, use only the step-level annotations, and apply text normalization with spaCy. CrossTask is an instructional video benchmark for zero-shot step localization. It contains approximately 4.8k instructional videos, covering 18 primary tasks and 65 related tasks. We follow the same evaluation set as \cite{li2024nasva, zanella2025baglm}, using videos from primary tasks.

\noindent \textbf{Metrics.} For HT-Step and CrossTask, we evaluate grounding performance using mAP@IoU following the HT-Step evaluation protocol. For Ego4D Goal-Step, we follow its official protocol and report Recall@1 at mIoU = 0.3. Unlike the Recall@1 (R@1) metric adopted in prior works~\cite{mavroudi2023vina, li2024nasva, zanella2025baglm}, which only verifies whether the top-ranked prediction falls within the annotated interval, our evaluation additionally requires sufficient temporal overlap between predicted and ground-truth segments. This yields a stricter, segment-level assessment that penalizes imprecise, noisy, or fragmented predictions, and thus constitutes a more appropriate measure of grounding quality. For completeness, we report R@1 results and discuss the differences between the two evaluation protocols in Appendix~\ref{app:eval_protocol}.


\noindent\textbf{Baselines.}
We compare \methodShort{} with several baselines.
First, among VSG methods, we consider the only two online approaches in
the literature, BAGLM~\cite{zanella2025baglm} and NaSVA-Online, the online variant
of NaSVA introduced in~\cite{zanella2025baglm}, together with the offline methods 
TAN~\cite{han2022tan} and NaSVA~\cite{li2024nasva}.
Second, we adapt dense video captioning (DVC) models, which predict temporal
boundaries together with free-form captions. Since no online DVC model releases
official code, we use the offline methods Vid2Seq~\cite{yang2023vid2seq},
VidChapters-7M~\cite{yang2023vidchapters} and TRACE~\cite{guo2025trace}, and adapt
each in two ways: the \textit{text} variant assigns every predicted caption to the
candidate step with the highest Sentence-BERT
similarity~\cite{reimers2019sentence}, while the \textit{boundary} variant discards
the captions and applies the same zero-shot event-to-step grounding as \methodShort{} to the predicted boundaries.
Third, we adapt UniTime~\cite{li2026universal}, a state-of-the-art offline video temporal grounding model trained on large-scale data,\footnote{Its training data includes datasets built from Ego4D narrations, so Ego4D Goal-Step validation videos may have been seen in pretraining.} which returns the temporal boundaries of a given textual description in the video. We adapt it by querying it once for each candidate step and taking the returned boundaries. Finally, we report the performance of an oracle that applies the same zero-shot event-to-step grounding as \methodShort{} to ground-truth temporal segments.

For VSG methods, we report numbers from~\cite{zanella2025baglm} on HT-Step, while on
Ego4D Goal-Step and CrossTask we reproduce all methods with their official code. For methods operating on fixed-length units, final merged segments
are obtained with a 1D blob detector, following~\cite{zanella2025baglm,
wang2023egoonly, li2024nasva}.

\noindent \textbf{Implementation Details.}
For fair comparison with prior work~\cite{zanella2025baglm}, each video is processed sequentially as a stream of non-overlapping 2-second segments sampled at 2 fps. Since $E_j$ is incrementally extended until the continuity criterion is violated, we uniformly sample 16 frames for the subsequent stages. 
In the Hierarchical Refinement stage, we set the entropy threshold to $\eta = 0.05$, use max aggregation over substeps, and refine at most $k = 10$ candidate steps, with the number of generated substeps ranging from 1 to 5. 
In all settings, adjacent segments sharing the same predicted step are merged, with the final score averaged over the constituent segments. All experiments are conducted on a single NVIDIA A100 GPU.

\subsection{Comparison with the State of the Art}
\label{sec:sota}

Table~\ref{tab:main-sota} compares \methodShort{} with the state of the art on HT-Step, CrossTask and Ego4D Goal-Step. Across all backbones and benchmarks, \methodShort{} consistently outperforms the corresponding BAGLM variant, establishing a new state of the art for online VSG. On HT-Step, with the same Qwen2.5-VL-7B backbone, \methodShort{} improves the average mAP from 7.70 to 14.68 over BAGLM. The gains are consistent across all IoU thresholds and are particularly pronounced under stricter localization criteria: at IoU$=0.7$, \methodShort{} improves from 1.70 to 5.74. Qwen3-VL-8B improves performance further, indicating that our method scales effectively with stronger backbones. The same behavior is observed on CrossTask and Ego4D Goal-Step, where Qwen2.5-VL-7B improves from 11.43 to 18.06 average mAP and from 13.09 to 19.51 R@1 at mIoU=0.3, respectively. \methodShort{} also surpasses the offline methods TAN and NaSVA on all three benchmarks. Notably, the oracle, which considers ground-truth temporal segments, remains far from perfect ($31.52$ and $42.35$ average mAP on HT-Step and CrossTask with Qwen3-VL-8B), confirming the difficulty of the task.

It is also interesting to evaluate the performance of other methods adapted to the VSG task. While DVC-based baselines remain clearly behind \methodShort{}, they provide useful insights into the sources of difficulty in performing online VSG. For all three DVC models, the
\textit{boundary} variant outperforms its \textit{text} counterpart by a large
margin (e.g., $9.01$ vs.\ $3.86$ average mAP for Vid2Seq on HT-Step). Since the two
variants share the same predicted boundaries and differ only in how segments are
matched to steps, the gap isolates the captions as the weak component: DVC models
do localize meaningful temporal units, but their free-form captions describe visual
appearance rather than step identity, and are therefore poor proxies for procedural
step descriptions. UniTime is the strongest baseline, reaching $12.27$ and $18.68$ average mAP on HT-Step and CrossTask, and $24.87$ R@1 at mIoU on Ego4D Goal-Step. The comparison is however unfavorable to HSD, as UniTime operates in the easier offline setting, it requires multiple forward passes (one for each candidate procedural step) and is trained on a large-scale grounding corpus covering the same instructional domain. 

\begin{figure}[t]
\centering
\includegraphics[width=\linewidth]{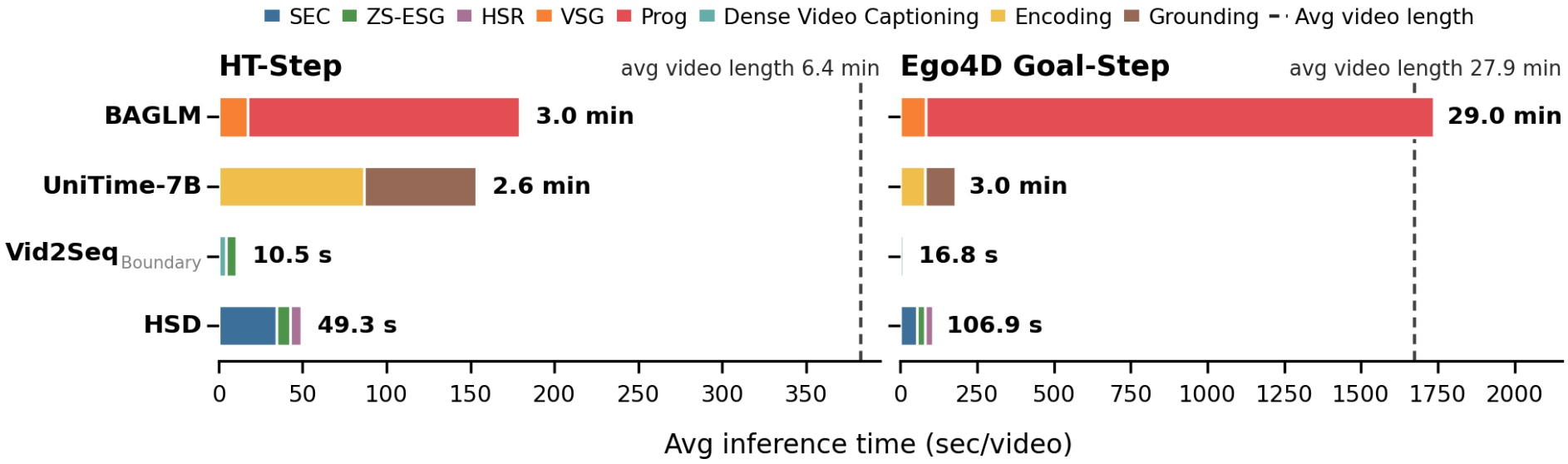}
\caption{
Inference-time comparison between BAGLM~\cite{zanella2025baglm}, UniTime-7B~\cite{li2026universal}, Vid2Seq\textsubscript{\textcolor{lightgraytext}{Boundary}}~\cite{yang2023vid2seq} and \methodShort{} on HT-Step and Ego4D Goal-Step using Qwen2.5-VL-7B.
}
\label{fig:inference_speed}
\end{figure}

\noindent \textbf{Inference speed comparison.} Beyond grounding accuracy, we analyze the inference cost of \methodShort{}
against BAGLM~\cite{zanella2025baglm}, the state-of-the-art OVSG method, and the best competitive baselines 
UniTime~\cite{li2026universal} and Vid2Seq~\cite{yang2023vid2seq}.
We use the same Qwen2.5-VL-7B backbone for BAGLM and \methodShort{} on a single
NVIDIA A100 64GB GPU, treating video loading as shared and reporting the average
runtime of the method-specific components (Figure~\ref{fig:inference_speed}):
Semantic Event Construction (SEC), Zero-Shot Event-to-Step Grounding (ZS-ESG)
and Hierarchical Step Refinement (HSR) for \methodShort{}, Visual Step Grounding
(VSG) and Progress Estimation (Prog) for BAGLM, and encoding and grounding for
UniTime.

On HT-Step, where videos last 6.4 minutes on average, \methodShort{} requires
49.3 seconds per video against 3.0 minutes for BAGLM and 2.6 minutes for
UniTime-7B, i.e. a 3.6$\times$ and 3.2$\times$ speedup. The gap widens on
Ego4D Goal-Step, with videos of 27.9 minutes: \methodShort{} takes 106.9 seconds,
whereas BAGLM requires 29.0 minutes, longer than the video itself, and UniTime
3.0 minutes, yielding a 16.3$\times$ and 1.7$\times$ speedup. This follows from
the different computational granularity: BAGLM grounds and estimates progress on
every fixed 2-second segment and UniTime needs one forward pass per candidate step,
whereas \methodShort{} applies ZS-ESG, and possibly HSR, to semantically coherent
events, avoiding redundant grounding and scaling better to long videos.
Vid2Seq$_\text{Boundary}$ is the only faster method, but it is offline and far less accurate
(Table~\ref{tab:main-sota}).

\begin{figure*}[t]
    \includegraphics[width=0.94\linewidth]{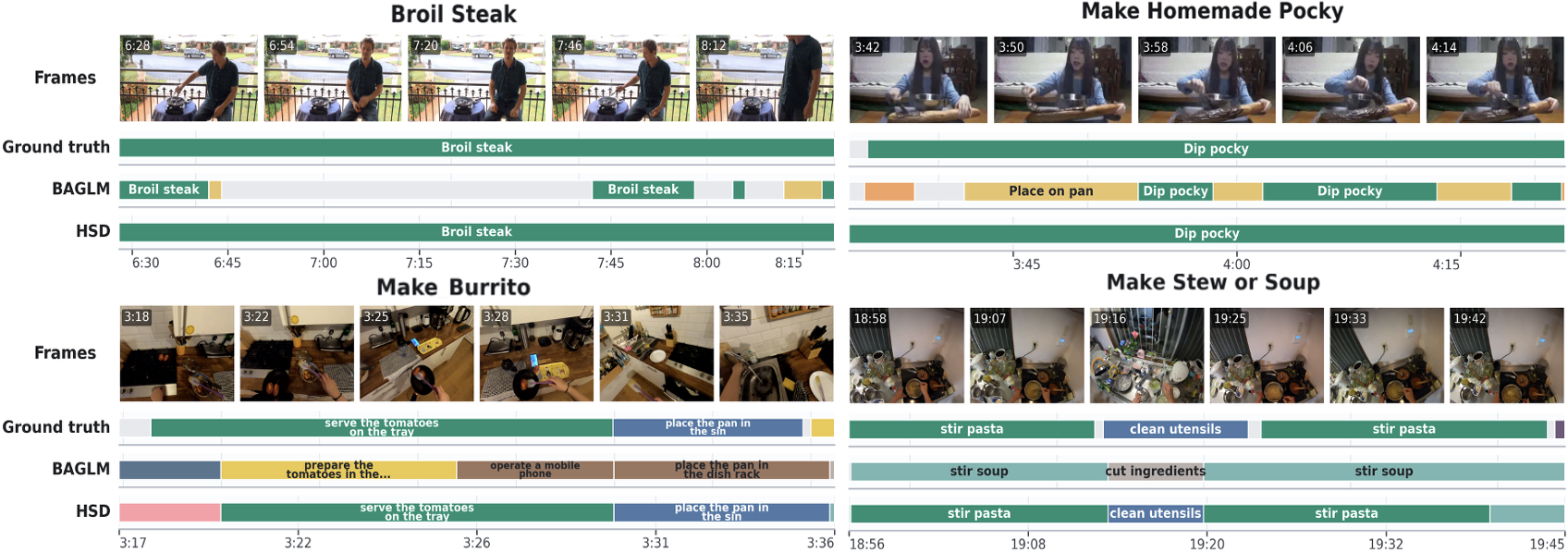}
    \caption{Qualitative comparison between \methodShort{} and BAGLM.}
    \label{fig:qualitative_lr}
\end{figure*}

\noindent \textbf{Qualitative comparison.}
Figure~\ref{fig:qualitative_lr} shows four qualitative comparisons between BAGLM~\cite{zanella2025baglm} and \methodShort{} using Qwen2.5-VL-7B. Across the \textit{Broil Steak}, \textit{Make Homemade Pocky}, and \textit{Make Burrito} examples, BAGLM produces fragmented or noisy predictions, likely due to noisy visual evidence in the fixed-duration segments. For instance, it interrupts the central \textit{broil steak} action with \textit{None}, confuses \textit{prepare tomato} with \textit{serve tomato}, and incorrectly predicts \textit{Place on pan} in the \textit{Make Homemade Pocky} video, where the pan in the scene may bias the prediction toward an incorrect step. By constructing semantically consistent events, \methodShort{} reduces the influence of such spurious cues and produces more coherent temporal localizations. In the \textit{Make Stew or Soup} example, \methodShort{} further benefits from its hierarchical refinement stage. Here, the generated substeps \textit{rinse the utensil} and \textit{move the pasta around with a spoon} provide more evidence for grounding the higher-level steps \textit{clean utensils} and \textit{stir pasta}, respectively, which would otherwise be confused with semantically similar steps.
These examples highlight how \methodShort{} improves temporal coherence and semantic discrimination in challenging procedural videos.

\subsection{Ablation studies}
\label{sec:ablation}
In this section, we analyze the main components and design choices of \methodShort{}. We first evaluate the individual and combined contributions of Semantic Event Construction and Hierarchical Step Refinement. We then study different refinement activation strategies and, finally, assess the substep score weighting and diverse aggregation strategies. Additional studies are provided in Appendix~\ref{app:hyperparameter-analysis}.

\noindent\textbf{Impact of Semantic Event Construction and Hierarchical Step Refinement.}
We first evaluate the contribution of the two main components of the proposed framework, \textit{i.e.}, \emph{OSEC} (Online Semantic Event Construction) and \emph{HSR} (Hierarchical Step Refinement), through an ablation study. Table~\ref{tab:components_analysis} reports the results using Qwen2.5-VL-7B evaluated on HT-Step and Ego4D Goal-Step. As a base variant, we directly ground procedural steps on fixed-length video segments using the same LMM, without semantic event construction or hierarchical refinement. Adding \emph{OSEC} alone already yields a clear improvement on both benchmarks, increasing performance on HT-Step from 18.10 to 23.65 at IoU$=0.3$, and on Ego4D Goal-Step from 8.08 to 17.48 R@1 at mIoU$=0.3$. This confirms that semantically coherent event segments online provides a much stronger temporal support for grounding than fixed-length segments. Adding \emph{HSR} further improves performance, with gains of +2.27 on HT-Step at IoU$=0.3$ and +2.03 R@1 at mIoU$=0.3$ on Ego4D Goal-Step. This shows that hierarchical substep reasoning brings additional fine-grained semantic evidence once predictions are grounded on more coherent event segments. Furthermore, applying \emph{HSR} alone also improves over the base method, indicating that the two components are individually useful and complementary, with the best performance obtained when both are used.

\begin{table}[t]
\centering
\scriptsize
\setlength{\tabcolsep}{4pt}
\renewcommand{\arraystretch}{1.1}
\caption{
Component ablation study on HT-Step and Ego4D Goal-Step using Qwen2.5-VL-7B.
}
\label{tab:components_analysis}
\resizebox{0.95\linewidth}{!}{%
\begin{tabular}{cc ccccc}
\toprule
\multirow{2}{*}{\textbf{OSEC}} 
& \multirow{2}{*}{\textbf{HSR}} 
& \multicolumn{4}{c}{\textbf{HT-Step mAP@IoU $\uparrow$}} 
& \multirow{2}{*}{%
  \raisebox{-3.0ex}{%
    \shortstack[c]{%
      \textbf{Ego4D Goal-Step}\\
      \textbf{R@1, IoU=0.3 $\uparrow$}
    }%
  }%
} \\
\cmidrule(lr){3-6}
& & \textbf{@0.3} & \textbf{@0.5} & \textbf{@0.7} & \textbf{@[0.3--0.7]} 
& \\
\midrule
\xmark & \xmark 
& 18.10 & 8.84 & 3.53 & 9.87 & 8.08 \\
\xmark & \textcolor{blue}{\checkmark} 
& 20.89 & 10.31 & 3.95 & 11.44 & 8.22 \\
\textcolor{orange}{\checkmark} & \xmark 
& 23.65 & 12.56 & 5.73 & 13.72 & 17.48 \\
\rowcolor{secgray}
\textcolor{orange}{\checkmark} & \textcolor{blue}{\checkmark} 
& \textbf{25.92} & \textbf{13.43} & \textbf{5.74} & \textbf{14.68} & \textbf{19.51} \\
\bottomrule
\end{tabular}%
}
\end{table}

\begin{table}[t]
\centering
\scriptsize
\setlength{\tabcolsep}{6pt}
\renewcommand{\arraystretch}{1.1}
\caption{
HSR activation strategy on HT-Step and Ego4D Goal-Step for Qwen2.5-VL-7B.
}
\label{tab:trigger_ablation}
\resizebox{\linewidth}{!}{%
\begin{tabular}{l ccccc}
\toprule
\multirow{2}{*}{\textbf{Trigger}} 
& \multicolumn{4}{c}{\textbf{HT-Step mAP@tIoU $\uparrow$}} 
& \multirow{2}{*}{%
  \raisebox{-3.0ex}{%
    \shortstack[c]{%
      \textbf{Ego4D Goal-Step}\\
      \textbf{R@1, IoU=0.3 $\uparrow$}
    }%
  }%
} \\
\cmidrule(lr){2-5}
& \textbf{@0.3} & \textbf{@0.5} & \textbf{@0.7} & \textbf{@[0.3--0.7]} 
& \\
\midrule
\textsc{None}    
& 23.65 & 12.56 & 5.73 & 13.72 & 17.48 \\
\textsc{Always}  
& 23.58 & 12.11 & 5.03 & 13.26 & 17.36 \\
\textsc{Random}  
& 18.05 & 9.05  & 3.85 & 10.64 & 17.09 \\
\rowcolor{secgray}
\textsc{Entropy} 
& \textbf{25.92} & \textbf{13.43} & \textbf{5.74} & \textbf{14.68} & \textbf{19.51} \\
\midrule
\textsc{Oracle}  
& 41.85 & 23.05 & 10.22 & 21.09 & 33.26 \\
\bottomrule
\end{tabular}%
}
\end{table}

\begin{table}[t]
\centering
\scriptsize
\setlength{\tabcolsep}{5.5pt}
\renewcommand{\arraystretch}{1.1}
\caption{
Ablation on Parent-Step Prior Weighting (PSPW) and Aggregation (Agg) on HT-Step and Ego4D Goal-Step using Qwen2.5-VL-7B.
}
\label{tab:aggregation_parent_weighting_ablation}
\resizebox{\linewidth}{!}{%
\begin{tabular}{ll ccccc}
\toprule
\multirow{2}{*}{\textbf{PSPW}} 
& \multirow{2}{*}{\textbf{Agg}} 
& \multicolumn{4}{c}{\textbf{HT-Step mAP@tIoU $\uparrow$}} 
& \multirow{2}{*}{%
  \raisebox{-3.0ex}{%
    \shortstack[c]{%
      \textbf{Ego4D Goal-Step}\\
      \textbf{R@1, IoU=0.3 $\uparrow$}
    }%
  }%
} \\
\cmidrule(lr){3-6}
& & \textbf{@0.3} & \textbf{@0.5} & \textbf{@0.7} & \textbf{@[0.3--0.7]} 
& \\
\midrule
\xmark & Sum & 24.50 & 12.79 & 5.35 & 14.21 & 17.44 \\
\checkmark & Sum & 25.03 & 12.93 & 5.59 & 14.52 & 19.02 \\
\xmark & Max & 25.05 & 13.30 & 5.52 & 14.62 & 17.53 \\
\rowcolor{secgray}
\checkmark & Max & \textbf{25.92} & \textbf{13.43} & \textbf{5.74} & \textbf{14.68} & \textbf{19.51} \\
\bottomrule
\end{tabular}%
}
\end{table}

\noindent\textbf{Substep refinement activation strategy.}
Table~\ref{tab:trigger_ablation} analyzes different strategies for deciding when to activate the Hierarchical Step Refinement stage using Qwen2.5-VL-7B variant evaluated on HT-Step and Ego4D Goal-Step. We compare four variants: \textsc{None}, which performs no refinement; \textsc{Always}, which applies refinement to every event segment; \textsc{Random}, which triggers refinement at random; \textsc{Entropy}, our proposed uncertainty-based criterion. We also consider an upper bound, \textit{i.e.}, \textsc{Oracle}, where refinement is activated only when the initial prediction is incorrect. Among these strategies, \textsc{Entropy} achieves the best results on both datasets, improving over \textsc{None} by +2.27 mAP@0.3 and +0.96 average mAP on HT-Step and +2.03 R@1 with mIoU$=0.3$ on Ego4D Goal-Step. It also outperforms \textsc{Always}, showing that refinement is most effective when applied selectively rather than indiscriminately.

\noindent\textbf{Analysis of Parent-Step Prior Weighting and Aggregation Strategies.}
We evaluate two key design choices in the Hierarchical Step Refinement stage: whether substep scores are reweighted by the probability of their parent step, as in Eq.~\eqref{eq:refined-normalization}, which we refer to as \emph{Parent-Step Prior Weighting} (PSPW), 
and how these scores are aggregated back to the step level, as in Eq.~\eqref{eq:substep_aggregation}, denoted \emph{Aggregation} (Agg). For Agg, we compare two strategies: \emph{Sum}, which accumulates evidence across all substeps, and \emph{Max}, which keeps the highest-scoring substep. Table~\ref{tab:aggregation_parent_weighting_ablation} reports the results using Qwen2.5-VL-7B on HT-Step and Ego4D Goal-Step. 
Across both datasets, PSPW consistently improves performance for both Agg strategies, highlighting the importance of preserving coarse step-level priors when incorporating fine-grained substep evidence. 
Additionally, the choice of Agg is also relevant: Max aggregation outperforms Sum aggregation in all settings, suggesting that relying on the most informative substep yields a stronger signal than aggregating potentially noisy evidence across all substeps.

\section{Conclusion}
\label{sec:conclusion}
We presented \methodShort{}, a training-free framework for online video step grounding that operates with a single frozen LMM. Unlike prior methods that ground procedural steps on fixed-length video segments, \methodShort{} incrementally constructs  adaptive temporal units that better align with the variable duration of real procedural steps, and decomposes candidate steps into finer-grained substeps to refine uncertain predictions addressing the granularity mismatch between coarse step descriptions and local visual evidence. Experiments on HT-Step, Ego4D Goal-Step and CrossTask demonstrate that \methodShort{} achieves state-of-the-art performance. We also compare VSG against adjacent settings, so far unexplored, and find that these perform poorly with procedural steps. The results highlight the importance of adapting both the temporal unit and the semantic granularity of steps descriptions in online VSG.
\clearpage
\paragraph{Acknowledgments.} This work was supported by the EU Horizon ELIAS (No. 101120237), ELLIOT (No. 101214398), TURING (No. 101215032) and FIS project GUIDANCE (No. FIS2023-03251) projects. The authors acknowledge the CINECA award under the ISCRA initiative for the availability of high performance computing resources and support.

{
    \small
    \bibliographystyle{ieeenat_fullname}
    \bibliography{main}

@String(CVPR= {IEEE Conf. Comput. Vis. Pattern Recog.})

@String(ICCV= {Int. Conf. Comput. Vis.})

@String(ECCV= {Eur. Conf. Comput. Vis.})

@String(BMVC= {Brit. Mach. Vis. Conf.})

@String(ICLR = {Int. Conf. Learn. Represent.})

@String(CVPR  = {CVPR})

@String(ICCV  = {ICCV})

@String(ECCV  = {ECCV})

@String(BMVC  =	{BMVC})

@String(ICLR  = {ICLR})

@INPROCEEDINGS{Zhukov2019,
  author={Zhukov, Dimitri and Alayrac, Jean-Baptiste and Cinbis, Ramazan Gokberk and Fouhey, David and Laptev, Ivan and Sivic, Josef},
  booktitle={CVPR}, 
  title={Cross-Task Weakly Supervised Learning From Instructional Videos}, 
  year={2019}
  }

@inproceedings{mavroudi2023vina,
  title={Learning to ground instructional articles in videos through narrations},
  author={Mavroudi, Effrosyni and Afouras, Triantafyllos and Torresani, Lorenzo},
  booktitle={ICCV},
  year={2023}
}

@inproceedings{chen2024mptva,
  title={Learning to localize actions in instructional videos with llm-based multi-pathway text-video alignment},
  author={Chen, Yuxiao and Li, Kai and Bao, Wentao and Patel, Deep and Kong, Yu and Min, Martin Renqiang and Metaxas, Dimitris N},
  booktitle={ECCV},
  year={2024},
}

@inproceedings{li2024nasva,
  title={Multi-sentence grounding for long-term instructional video},
  author={Li, Zeqian and Chen, Qirui and Han, Tengda and Zhang, Ya and Wang, Yanfeng and Xie, Weidi},
  booktitle={ECCV},
  year={2024},
}

@inproceedings{zanella2025baglm,
  title={Training-free Online Video Step Grounding},
  author={Zanella, Luca and Mancini, Massimiliano and Wang, Yiming and Tonioni, Alessio and Ricci, Elisa},
  booktitle={NeurIPS},
  year={2025},
}

@article{afouras2023htstep,
  title={Ht-step: Aligning instructional articles with how-to videos},
  author={Afouras, Triantafyllos and Mavroudi, Effrosyni and Nagarajan, Tushar and Wang, Huiyu and Torresani, Lorenzo},
  journal={NeurIPS},
  year={2023}
}

@article{song2023ego4dgs,
  title={Ego4d goal-step: Toward hierarchical understanding of procedural activities},
  author={Song, Yale and Byrne, Eugene and Nagarajan, Tushar and Wang, Huiyu and Martin, Miguel and Torresani, Lorenzo},
  journal={NeurIPS},
  year={2023}
}

@InProceedings{Grauman2022ego4d,
    author    = {Grauman, Kristen and Westbury, Andrew and Byrne, Eugene and Chavis, Zachary and Furnari, Antonino and Girdhar, Rohit and Hamburger, Jackson and Jiang, Hao and Liu, Miao and Liu, Xingyu and Martin, Miguel and Nagarajan, Tushar and Radosavovic, Ilija and Ramakrishnan, Santhosh Kumar and Ryan, Fiona and Sharma, Jayant and Wray, Michael and Xu, Mengmeng and Xu, Eric Zhongcong and Zhao, Chen and Bansal, Siddhant and Batra, Dhruv and Cartillier, Vincent and Crane, Sean and Do, Tien and Doulaty, Morrie and Erapalli, Akshay and Feichtenhofer, Christoph and Fragomeni, Adriano and Fu, Qichen and Gebreselasie, Abrham and Gonz\'alez, Cristina and Hillis, James and Huang, Xuhua and Huang, Yifei and Jia, Wenqi and Khoo, Weslie and Kol\'a\v{r}, J\'achym and Kottur, Satwik and Kumar, Anurag and Landini, Federico and Li, Chao and Li, Yanghao and Li, Zhenqiang and Mangalam, Karttikeya and Modhugu, Raghava and Munro, Jonathan and Murrell, Tullie and Nishiyasu, Takumi and Price, Will and Ruiz, Paola and Ramazanova, Merey and Sari, Leda and Somasundaram, Kiran and Southerland, Audrey and Sugano, Yusuke and Tao, Ruijie and Vo, Minh and Wang, Yuchen and Wu, Xindi and Yagi, Takuma and Zhao, Ziwei and Zhu, Yunyi and Arbel\'aez, Pablo and Crandall, David and Damen, Dima and Farinella, Giovanni Maria and Fuegen, Christian and Ghanem, Bernard and Ithapu, Vamsi Krishna and Jawahar, C. V. and Joo, Hanbyul and Kitani, Kris and Li, Haizhou and Newcombe, Richard and Oliva, Aude and Park, Hyun Soo and Rehg, James M. and Sato, Yoichi and Shi, Jianbo and Shou, Mike Zheng and Torralba, Antonio and Torresani, Lorenzo and Yan, Mingfei and Malik, Jitendra},
    title     = {Ego4D: Around the World in 3,000 Hours of Egocentric Video},
    booktitle = {CVPR},
    year      = {2022},
}

@article{shen2024protas,
  author = {Y.~Shen and E.~Elhamifar},
  title = {Progress-Aware Online Action Segmentation for Egocentric Procedural Task Videos},
  journal = {CVPR},
  year = {2024}}

@inproceedings{jung2025ogebd,
  author    = {Jung, Hyungrok and Kim, Daneul and Lim, Seunggyun and Son, Jeany and Choi, Jonghyun},
  title     = {Online Generic Event Boundary Detection},
  booktitle = {ICCV},
  year      = {2025},
}

@inproceedings{mounir2023streamer,
  title={STREAMER: Streaming Representation Learning and Event Segmentation in a Hierarchical Manner},
  author={Mounir, Ramy and Vijayaraghavan, Sujal and Sarkar, Sudeep},
  booktitle={NeurIPS},
  year={2023}
}

@InProceedings{liberatori2024t3al,
    author    = {Liberatori, Benedetta and Conti, Alessandro and Rota, Paolo and Wang, Yiming and Ricci, Elisa},
    title     = {Test-Time Zero-Shot Temporal Action Localization},
    booktitle = {CVPR},
    year      = {2024},
}

@article{han2025freezad,
  title={Training-Free Zero-Shot Temporal Action Detection with Vision-Language Models},
  author={Han, Chaolei and Wang, Hongsong and Kuang, Jidong and Zhang, Lei and Gui, Jie},
  journal={arXiv},
  year={2025}
}

@article{zhang2021vslnet,
  title={Natural language video localization: A revisit in span-based question answering framework},
  author={Zhang, Hao and Sun, Aixin and Jing, Wei and Zhen, Liangli and Zhou, Joey Tianyi and Goh, Rick Siow Mong},
  journal={IEEE TPAMI},
  year={2021},
}

@misc{bai2025qwen25vltechnicalreport,
      title={Qwen2.5-VL Technical Report}, 
      author={Shuai Bai and Keqin Chen and Xuejing Liu and Jialin Wang and Wenbin Ge and Sibo Song and Kai Dang and Peng Wang and Shijie Wang and Jun Tang and Humen Zhong and Yuanzhi Zhu and Mingkun Yang and Zhaohai Li and Jianqiang Wan and Pengfei Wang and Wei Ding and Zheren Fu and Yiheng Xu and Jiabo Ye and Xi Zhang and Tianbao Xie and Zesen Cheng and Hang Zhang and Zhibo Yang and Haiyang Xu and Junyang Lin},
      year={2025},
      archivePrefix={arXiv}, 
}

@misc{bai2025qwen3vltechnicalreport,
      title={Qwen3-VL Technical Report}, 
      author={Shuai Bai and Yuxuan Cai and Ruizhe Chen and Keqin Chen and Xionghui Chen and Zesen Cheng and Lianghao Deng and Wei Ding and Chang Gao and Chunjiang Ge and Wenbin Ge and Zhifang Guo and Qidong Huang and Jie Huang and Fei Huang and Binyuan Hui and Shutong Jiang and Zhaohai Li and Mingsheng Li and Mei Li and Kaixin Li and Zicheng Lin and Junyang Lin and Xuejing Liu and Jiawei Liu and Chenglong Liu and Yang Liu and Dayiheng Liu and Shixuan Liu and Dunjie Lu and Ruilin Luo and Chenxu Lv and Rui Men and Lingchen Meng and Xuancheng Ren and Xingzhang Ren and Sibo Song and Yuchong Sun and Jun Tang and Jianhong Tu and Jianqiang Wan and Peng Wang and Pengfei Wang and Qiuyue Wang and Yuxuan Wang and Tianbao Xie and Yiheng Xu and Haiyang Xu and Jin Xu and Zhibo Yang and Mingkun Yang and Jianxin Yang and An Yang and Bowen Yu and Fei Zhang and Hang Zhang and Xi Zhang and Bo Zheng and Humen Zhong and Jingren Zhou and Fan Zhou and Jing Zhou and Yuanzhi Zhu and Ke Zhu},
      year={2025},
      booktitle={arXiv},
}

@inproceedings{kang2021cagil,
  title={Cag-qil: Context-aware actionness grouping via q imitation learning for online temporal action localization},
  author={Kang, Hyolim and Kim, Kyungmin and Ko, Yumin and Kim, Seon Joo},
  booktitle={ICCV},
  year={2021}
}

@inproceedings{kim2022sliding,
  title={A Sliding Window Scheme for Online Temporal Action Localization},
  author={Kim, Young Hwi and Kang, Hyolim and Kim, Seon Joo},
  booktitle={ECCV},
  year={2022},
}

@inproceedings{zhong2024ontas,
 author = {Zhong, Qing and Ding, Guodong and Yao, Angela},
 booktitle = {NeurIPS},
 title = {OnlineTAS: An Online Baseline for Temporal Action Segmentation},
 year = {2024}
}

@InProceedings{han2022tan,
  title={Temporal Alignment Network for long-term Video},  
  author={Tengda Han and Weidi Xie and Andrew Zisserman},  
  booktitle={CVPR},  
  year={2022}}

@misc{chen2025internvl25,
      title={Expanding Performance Boundaries of Open-Source Multimodal Models with Model, Data, and Test-Time Scaling}, 
      author={Zhe Chen and Weiyun Wang and Yue Cao and Yangzhou Liu and Zhangwei Gao and Erfei Cui and Jinguo Zhu and Shenglong Ye and Hao Tian and Zhaoyang Liu and Lixin Gu and Xuehui Wang and Qingyun Li and Yiming Ren and Zixuan Chen and Jiapeng Luo and Jiahao Wang and Tan Jiang and Bo Wang and Conghui He and Botian Shi and Xingcheng Zhang and Han Lv and Yi Wang and Wenqi Shao and Pei Chu and Zhongying Tu and Tong He and Zhiyong Wu and Huipeng Deng and Jiaye Ge and Kai Chen and Kaipeng Zhang and Limin Wang and Min Dou and Lewei Lu and Xizhou Zhu and Tong Lu and Dahua Lin and Yu Qiao and Jifeng Dai and Wenhai Wang},
      year={2024},
      booktitle={arXiv}
      }

@article{wang2025internvl3,
  title={Internvl3. 5: Advancing open-source multimodal models in versatility, reasoning, and efficiency},
  author={Wang, Weiyun and Gao, Zhangwei and Gu, Lixin and Pu, Hengjun and Cui, Long and Wei, Xingguang and Liu, Zhaoyang and Jing, Linglin and Ye, Shenglong and Shao, Jie and others},
  journal={arXiv},
  year={2025}
}

@inproceedings{reza2024hat,
  title={HAT: History-Augmented Anchor Transformer for Online Temporal Action Localization},
  author={Reza, Sakib and Zhang, Yuexi and Moghaddam, Mohsen and Camps, Octavia},
  booktitle={ECCV},
  year={2024},
}

@inproceedings{song2024matr,
    title={Online Temporal Action Localization with Memory-Augmented Transformer},
    author={Song, Youngkil and Kim, Dongkeun and Cho, Minsu and Kwak, Suha},
    booktitle={ECCV},
    year={2024},
}

@inproceedings{miech2019howto100m,
  title={Howto100m: Learning a text-video embedding by watching hundred million narrated video clips},
  author={Miech, Antoine and Zhukov, Dimitri and Alayrac, Jean-Baptiste and Tapaswi, Makarand and Laptev, Ivan and Sivic, Josef},
  booktitle={ICCV},
  year={2019}
}

@inproceedings{xu2021videoclip,
  title={Videoclip: Contrastive pre-training for zero-shot video-text understanding},
  author={Xu, Hu and Ghosh, Gargi and Huang, Po-Yao and Okhonko, Dmytro and Aghajanyan, Armen and Metze, Florian and Zettlemoyer, Luke and Feichtenhofer, Christoph},
  booktitle={EMNLP},
  year={2021}
}

@inproceedings{chen2021mcn,
  title={Multimodal clustering networks for self-supervised learning from unlabeled videos},
  author={Brian Chen and Andrew Rouditchenko and Kevin Duarte and Hilde Kuehne and Samuel Thomas and Angie Boggust and Rameswar Panda and Brian Kingsbury and Rogerio Feris and David Harwath and James Glass and Michael Picheny and Shih-Fu Chang},
  booktitle={ICCV},
  year={2021}
}

@inproceedings{shen2021dwsa,
  title={Learning to segment actions from visual and language instructions via differentiable weak sequence alignment},
  author={Shen, Yuhan and Wang, Lu and Elhamifar, Ehsan},
  booktitle={CVPR},
  year={2021}
}

@inproceedings{miech2020milnce,
  title={End-to-end learning of visual representations from uncurated instructional videos},
  author={Miech, Antoine and Alayrac, Jean-Baptiste and Smaira, Lucas and Laptev, Ivan and Sivic, Josef and Zisserman, Andrew},
  booktitle={CVPR},
  year={2020}
}

@inproceedings{ko2022vttwins,
  title={Video-text representation learning via differentiable weak temporal alignment},
  author={Ko, Dohwan and Choi, Joonmyung and Ko, Juyeon and Noh, Shinyeong and On, Kyoung-Woon and Kim, Eun-Sol and Kim, Hyunwoo J},
  booktitle={CVPR},
  year={2022}
}

@article{luo2020univl,
  title={Univl: A unified video and language pre-training model for multimodal understanding and generation},
  author={Luo, Huaishao and Ji, Lei and Shi, Botian and Huang, Haoyang and Duan, Nan and Li, Tianrui and Li, Jason and Bharti, Taroon and Zhou, Ming},
  journal={arXiv},
  year={2020}
}

@INPROCEEDINGS {wang2023egoonly,
author = { Wang, Huiyu and Singh, Mitesh Kumar and Torresani, Lorenzo },
booktitle = {ICCV},
title = {Ego-Only: Egocentric Action Detection without Exocentric Transferring},
year = {2023},
}

@inproceedings{yang2023vid2seq,
title = {Vid2Seq: Large-Scale Pretraining of a Visual Language Model for Dense Video Captioning},
author={Antoine Yang and Arsha Nagrani and Paul Hongsuck Seo and Antoine Miech and Jordi Pont-Tuset and Ivan Laptev and Josef Sivic and Cordelia Schmid},
booktitle={CVPR},
year = {2023}}

@article{lei2021detecting,
  title={Detecting moments and highlights in videos via natural language queries},
  author={Lei, Jie and Berg, Tamara L and Bansal, Mohit},
  journal={NeurIPS},
  year={2021}
}

@article{yang2023vidchapters,
  title={Vidchapters-7m: Video chapters at scale},
  author={Yang, Antoine and Nagrani, Arsha and Laptev, Ivan and Sivic, Josef and Schmid, Cordelia},
  journal={NeurIPS},
  year={2023}
}

@inproceedings{guo2025trace,
  title={Trace: Temporal grounding video llm via causal event modeling},
  author={Guo, Yongxin and Liu, Jingyu and Li, Mingda and Liu, Qingbin and Chen, Xi and Tang, Xiaoying},
  booktitle={ICLR},
  year={2025}
}

@inproceedings{farha2019ms,
  title={Ms-tcn: Multi-stage temporal convolutional network for action segmentation},
  author={Farha, Yazan Abu and Gall, Juergen},
  booktitle={CVPR},
  year={2019}
}

@article{yi2021asformer,
  title={Asformer: Transformer for action segmentation},
  author={Yi, Fangqiu and Wen, Hongyu and Jiang, Tingting},
  journal={BMVC},
  year={2021}
}

@inproceedings{zhang2022actionformer,
  title={Actionformer: Localizing moments of actions with transformers},
  author={Zhang, Chen-Lin and Wu, Jianxin and Li, Yin},
  booktitle={European Conference on Computer Vision},
  year={2022},
}

@inproceedings{shi2023tridet,
  title={Tridet: Temporal action detection with relative boundary modeling},
  author={Shi, Dingfeng and Zhong, Yujie and Cao, Qiong and Ma, Lin and Li, Jia and Tao, Dacheng},
  booktitle={CVPR},
  year={2023}
}

@inproceedings{gan2023tsgsv,
  title={Temporal Sentence Grounding in Streaming Videos},
  author={Gan, Tian and Wang, Xiao and Sun, Yan and Wu, Jianlong and Guo, Qingpei and Nie, Liqiang},
  booktitle={ACM MM},
  year={2023}
}

@inproceedings{zhou2024streamingdvc,
  title={Streaming Dense Video Captioning},
  author={Zhou, Xingyi and Arnab, Anurag and Buch, Shyamal and Yan, Shen and Myers, Austin and Xiong, Xuehan and Nagrani, Arsha and Schmid, Cordelia},
  booktitle={CVPR},
  year={2024}
}

@inproceedings{nag2022stale,
  title={Zero-Shot Temporal Action Detection via Vision-Language Prompting},
  author={Nag, Sauradip and Zhu, Xiatian and Song, Yi-Zhe and Xiang, Tao},
  booktitle={ECCV},
  year={2022},
}

@inproceedings{pramanick2025enrich,
  title={Enrich and detect: Video temporal grounding with multimodal llms},
  author={Pramanick, Shraman and Mavroudi, Effrosyni and Song, Yale and Chellappa, Rama and Torresani, Lorenzo and Afouras, Triantafyllos},
  booktitle={ICCV},
  year={2025},
}

@inproceedings{ren2024timechat,
  title={Timechat: A time-sensitive multimodal large language model for long video understanding},
  author={Ren, Shuhuai and Yao, Linli and Li, Shicheng and Sun, Xu and Hou, Lu},
  booktitle={CVPR},
  year={2024},
}

@article{li2026universal,
  title={Universal video temporal grounding with generative multi-modal large language models},
  author={Li, Zeqian and Di, Shangzhe and Zhai, Zhonghua and Huang, Weilin and Wang, Yanfeng and Xie, Weidi},
  journal={NeurIPS},
  year={2025}
}

@inproceedings{lin2023univtg,
  title={Univtg: Towards unified video-language temporal grounding},
  author={Lin, Kevin Qinghong and Zhang, Pengchuan and Chen, Joya and Pramanick, Shraman and Gao, Difei and Wang, Alex Jinpeng and Yan, Rui and Shou, Mike Zheng},
  booktitle={ICCV},
  year={2023},
}

@inproceedings{zheng2025hierarchical,
  title={Hierarchical event memory for accurate and low-latency online video temporal grounding},
  author={Zheng, Minghang and Peng, Yuxin and Sun, Benyuan and Yang, Yi and Liu, Yang},
  booktitle={ICCV},
  year={2025},
}

@inproceedings{reimers2019sentence,
  title={Sentence-bert: Sentence embeddings using siamese bert-networks},
  author={Reimers, Nils and Gurevych, Iryna},
  booktitle={EMNLP-IJCNLP},
  year={2019}
}

@inproceedings{shou2021generic,
  title={Generic event boundary detection: A benchmark for event segmentation},
  author={Shou, Mike Zheng and Lei, Stan Weixian and Wang, Weiyao and Ghadiyaram, Deepti and Feiszli, Matt},
  booktitle={ICCV},
  year={2021},
}

@article{oquab2024dinov2,
  title   = {{DINOv2}: Learning Robust Visual Features without Supervision},
  author  = {Oquab, Maxime and Darcet, Timoth{\'e}e and Moutakanni, Th{\'e}o and Vo, Huy and Szafraniec, Marc and Khalidov, Vasil and Fernandez, Pierre and Haziza, Daniel and Massa, Francisco and El-Nouby, Alaaeldin and others},
  journal = {TMLR},
  year    = {2024}
}
}

\clearpage
\section{Technical appendices and supplementary material}

\subsection{Evaluation protocol details}
\label{app:eval_protocol}

\paragraph{HT-Step and CrossTask evaluation.}
For HT-Step~\cite{afouras2023htstep} and CrossTask~\cite{Zhukov2019}, we follow the official article-grounding protocol and report mean Average Precision over temporal IoU thresholds. A predicted segment is correct if its step label matches the ground-truth step and its temporal IoU exceeds the threshold. We report mAP at temporal IoU thresholds $\{0.3, 0.5, 0.7\}$, as well as the average mAP over thresholds $\{0.3, 0.4, 0.5, 0.6, 0.7\}$, denoted as $\mathrm{mAP@[0.3\text{--}0.7]}$.

AP is computed separately for each activity by pooling all predictions and ground-truth intervals from videos of that activity; the final mAP is the average across activities. This metric is well suited to HT-Step and CrossTask because it evaluates both step identification and temporal extent, while also penalizing false positives on non-groundable article steps. In contrast, recall-based metrics ignore such false positives, and frame-wise metrics do not assess segment-level localization quality.

\noindent\textbf{Ego4D Goal-Step evaluation.}
For Ego4D Goal-Step~\cite{song2023ego4dgs}, we follow official evaluation protocol, reporting Recall@1 at $\mathrm{mIoU}=0.3$. A prediction is counted as correct if the top-ranked predicted segment for a step overlaps with the ground-truth interval with temporal IoU greater than $0.3$. As in the main paper, we evaluate only at the step level and apply the same text normalization procedure used by prior work~\cite{zanella2025baglm}, including lowercasing, lemmatization, and plural canonicalization.

\noindent\textbf{Relation to Recall@1.}
For completeness, we also report the results with Recall@1 in Appendix~\ref{app:add_anal_r1}. Recall@1 is useful for comparison with prior step-grounding methods, but it is less suitable compared to the official metric. Specifically, it only checks whether the highest-scoring temporal prediction falls inside the ground-truth interval, without evaluating the predicted segment as a whole. Therefore, a method can obtain high R@1 as long as one top-scoring frame lies within the ground-truth interval, even if the remaining predictions are fragmented or  imprecise. Moreover, R@1 does not penalize false positives on non-groundable article steps. For these reasons, we do not use R@1 as our primary metric. R@1 is also not fully aligned with our prediction format: our method produces variable-length segments and assigns the same score to all frames within each predicted segment, rather than producing frame-level scores. As a result, multiple frames may share the highest score. When computing R@1, we therefore count a prediction as correct if any of the top-scoring frames falls within the ground-truth interval.

\subsection{Additional analysis using Recall@1}
\label{app:add_anal_r1}
Table~\ref{tab:htstep-ego4d} reports additional results under the Recall@1 protocol commonly used in prior work. Although our method is designed for segment-level grounding, it remains competitive under this point-based metric and achieves the best performance among online methods on both HT-Step and Ego4D Goal-Step. In particular, it improves over BaGLM by +1.5 R@1 on HT-Step and +1.1 R@1 on Ego4D Goal-Step. For a fair comparison, we report both BaGLM and our method with the same Qwen2.5-VL-7B base model.

\begin{table*}[t]
\centering
\caption{
Performance comparison on CrossTask, HT-Step, and Ego4D Goal-Step.}
\label{tab:main-results}
\setlength{\tabcolsep}{5pt}
\renewcommand{\arraystretch}{1.12}

\begin{minipage}[t]{0.43\textwidth}
\centering
\captionof{table}{CrossTask}
\label{tab:crosstask}
\resizebox{\linewidth}{!}{%
\begin{tabular}{l c}
\toprule
\textbf{Method} & \textbf{Avg. R@1 $\uparrow$} \\
\midrule
\rowcolor{secblue}
\multicolumn{2}{l}{\textit{Offline methods}} \\
Zhukov \textit{et al.}~\cite{Zhukov2019} & 22.4 \\
HT100M~\cite{miech2019howto100m} & 33.6 \\
VideoCLIP~\cite{xu2021videoclip} & 33.9 \\
MCN~\cite{chen2021mcn} & 35.1 \\
DWSA~\cite{shen2021dwsa} & 35.3 \\
MIL-NCE~\cite{miech2020milnce} & 40.5 \\
VT-TWINS~\cite{ko2022vttwins} & 40.7 \\
UniVL~\cite{luo2020univl} & 42.0 \\
VINA~\cite{mavroudi2023vina} & 44.8 \\
NASVA~\cite{li2024nasva} & 46.7 \\
MPTVA~\cite{chen2024mptva} & 47.9 \\
\midrule
\rowcolor{secgreen}
\multicolumn{2}{l}{\textit{Online methods}} \\
BAGLM$_{\textcolor{gray}{\scriptsize\text{Qwen2.5-VL-7B}}}$~\cite{zanella2025baglm} & 53.3 \\
\textbf{Ours}$_{\textcolor{gray}{\scriptsize\text{Qwen2.5-VL-7B}}}$ & \textbf{53.6} \\
\bottomrule
\end{tabular}%
}
\end{minipage}
\hfill
\begin{minipage}[t]{0.47\textwidth}
\centering
\captionof{table}{HT-Step and Ego4D Goal-Step}
\label{tab:htstep-ego4d}
\resizebox{\linewidth}{!}{%
\begin{tabular}{l c c}
\toprule
\multirow{2}{*}{\textbf{Method}} 
& \textbf{HT-Step} 
& \textbf{Ego4D} \\
& \textbf{R@1 $\uparrow$} 
& \textbf{R@1 $\uparrow$} \\
\midrule
\rowcolor{secblue}
\multicolumn{3}{l}{\textit{Offline methods}} \\
VSLNet~\cite{zhang2021vslnet} & -- & 24.3 \\
TAN*~\cite{han2022tan} & 30.7 & -- \\
VINA~\cite{mavroudi2023vina} & 39.1 & -- \\
NaSVA~\cite{li2024nasva} & 53.1 & 29.1 \\
\midrule
\rowcolor{secgreen}
\multicolumn{3}{l}{\textit{Online methods}} \\
NaSVA-Online~\cite{li2024nasva} & 46.1 & 24.2 \\
BaGLM$_{\textcolor{gray}{\scriptsize\text{Qwen2.5-VL-7B}}}$~\cite{zanella2025baglm} & 54.3 & 36.3 \\
\textbf{Ours}$_{\textcolor{gray}{\scriptsize\text{Qwen2.5-VL-7B}}}$ & \textbf{55.8} & \textbf{37.4} \\
\bottomrule
\end{tabular}%
}
\end{minipage}

\vspace{-0.5em}
\end{table*}

\subsection{Hyperparameter Analysis}
\label{app:hyperparameter-analysis}

We analyze the sensitivity of \methodShort{} to the two main hyperparameters introduced by the Hierarchical Step Refinement module: the entropy threshold $\eta$, which controls whether refinement is activated, and the number of top-$k$ candidate steps considered during refinement. Experiments are conducted on Ego4D Goal-Step using Qwen2.5-VL-7B as the frozen LMM, and performance is reported in terms of R@1 at IoU threshold 0.3.

\noindent\textbf{Hierarchical Step Refinement activation threshold.}
The Hierarchical Step Refinement module is applied when the initial event-to-step prediction is uncertain. Specifically, after obtaining the coarse probability distribution $p_j$ over the candidate procedural steps, we compute its entropy $H_j$ and activate refinement when $H_j \geq \eta$. In this case, the model decomposes the most plausible high-level steps into finer-grained substeps, re-evaluates the event over this substep pool, and aggregates the resulting scores back to the original step space. Thus, $\eta$ determines how often this additional refinement stage is invoked.

Table~\ref{tab:ego4dgs_entropy_ablation} reports the effect of varying the entropy threshold $\eta \in \{0.05, 0.10, 0.15, 0.20\}$ while keeping $k=10$ fixed. Smaller values of $\eta$ activate refinement on a larger fraction of events, while larger values restrict refinement to only the most uncertain predictions. The results show that performance remains relatively stable across all tested thresholds, with R@1 at mIoU$=0.3$ varying only within a narrow range. This suggests that \methodShort{} is not highly sensitive to the exact value of $\eta$. Nevertheless, the best result is obtained with $\eta=0.05$, indicating that applying refinement more broadly is beneficial.

\noindent\textbf{Number of top-$k$ plausible step candidates.}
Once refinement is activated, \methodShort{} does not decompose all candidate steps. Instead, it first selects the top-$k$ most likely steps according to the initial distribution $p_j$, and only these candidates are expanded into substeps. This design keeps the refinement stage computationally tractable while focusing the additional reasoning on the most plausible hypotheses. The value of $M$ therefore controls the breadth of the refinement pool.

Table~\ref{tab:ego4dgs_topk_ablation} reports the effect of varying the number of candidate steps used during refinement, with $k \in \{4, 6, 8, 10\}$ and $\eta=0.05$ fixed. Increasing $k$ enlarges the set of coarse step hypotheses that are decomposed into substeps. As a result, the refinement stage has a higher chance of including the correct step among its candidate parents, at the cost of evaluating a larger substep pool. Performance improves when moving from $k=4$ to larger values, confirming that considering too few candidates can prematurely discard the correct step before refinement. The best performance is achieved with $k=10$, which suggests that a broader refinement pool provides useful additional coverage while remaining computationally manageable. Overall, these results indicate that \methodShort{} benefits from using hierarchical refinement over a sufficiently diverse set of plausible step hypotheses.

\begin{table*}[t]
\centering
\small
\setlength{\tabcolsep}{8pt}
\begin{minipage}{0.45\linewidth}
\centering
\begin{tabular}{cc}
\toprule
$\eta$ & R@1, IoU=0.3 \\
\midrule
0.05 & 0.1951 \\
0.10 & 0.1916 \\
0.15 & 0.1920 \\
0.20 & 0.1933 \\
\bottomrule
\end{tabular}
\caption{Sensitivity to the entropy threshold $\eta$ on Ego4D Goal-Step with $K{=}10$ and Qwen2.5-VL-7B. Lower values activate refinement for a larger fraction of events.}
\label{tab:ego4dgs_entropy_ablation}
\end{minipage}
\hfill
\begin{minipage}{0.45\linewidth}
\centering
\begin{tabular}{cc}
\toprule
$k$ & R@1, IoU=0.3 \\
\midrule
4  & 0.1867 \\
6  & 0.1920 \\
8  & 0.1916 \\
10 & 0.1951 \\
\bottomrule
\end{tabular}
\caption{Sensitivity to the number of top-$k$ candidate steps used for hierarchical refinement on Ego4D Goal-Step with $\eta{=}0.05$ and Qwen2.5-VL-7B.}
\label{tab:ego4dgs_topk_ablation}
\end{minipage}
\end{table*}

\subsection{Comparison with alternative temporal units.}
\label{app:osec_alternatives}
We compare OSEC with simpler ways of building temporal units, keeping the rest of the pipeline unchanged (Qwen2.5-VL-7B, without HSR). We consider longer fixed windows, boundaries detected from visual feature changes, and the boundaries predicted by Vid2Seq~\cite{yang2023vid2seq}. For feature changes, we encode each 1-second segment with DINOv2~\cite{oquab2024dinov2} and start a new event when its similarity to the current event drops below a threshold $\tau$. Table~\ref{tab:osec_alternatives} reports the results on HT-Step. Longer windows reach OSEC only at IoU$=0.3$, while feature-change boundaries are better than fixed windows at IoU$=0.7$ but remain below OSEC on average. Moreover, both require tuning a window length or a threshold, while OSEC does not.

\begin{table}[t]
\centering
\small
\setlength{\tabcolsep}{3.5pt}
\caption{Alternatives to Online Semantic Event Construction (OSEC) on HT-Step with Qwen2.5-VL-7B and no Hierarchical Step Refinement. All rows share the same zero-shot event-to-step grounding and merging; only the temporal unit changes. \#Units: average number of temporal units per video. Avg.: mAP averaged over tIoU $\in\{0.3,0.4,0.5,0.6,0.7\}$. $^\dagger$Offline.}
\label{tab:osec_alternatives}
\begin{tabular}{l r cccc}
\toprule
\textbf{Temporal unit} & \textbf{\#Units} & \textbf{@0.3} & \textbf{@0.5} & \textbf{@0.7} & \textbf{Avg.} \\
\midrule
\multicolumn{6}{l}{\textit{Fixed windows}} \\
2\,s (Table~\ref{tab:components_analysis}) & 190.7 & 18.10 & 8.84 & 3.53 & 9.87 \\
4\,s & 95.1 & 22.96 & 10.63 & 3.80 & 12.05 \\
8\,s & 47.3 & \textbf{23.81} & 10.47 & 3.12 & 11.87 \\
\midrule
\multicolumn{6}{l}{\textit{Feature-change boundaries (DINOv2)}} \\
$\tau=0.6$ & 19.9 & 17.79 & 9.59 & 4.59 & 10.49 \\
$\tau=0.7$ & 26.5 & 19.47 & 10.74 & 5.01 & 11.48 \\
$\tau=0.8$ & 36.6 & 21.74 & 12.05 & 5.56 & 12.72 \\
\midrule
\multicolumn{6}{l}{\textit{Learned boundaries}} \\
Vid2Seq$^\dagger$ (Table~\ref{tab:main-sota}) & 27.7 & 19.03 & 7.21 & 2.43 & 9.01 \\
\midrule
\rowcolor{secgray}
OSEC (ours) & 60.6 & 23.65 & \textbf{12.56} & \textbf{5.73} & \textbf{13.72} \\
\bottomrule
\end{tabular}
\end{table}

\subsection{Additional Method Details}
\label{app:method_details}
This section provides implementation and procedural details that complement the method description in Section~\ref{sec:method}. The exact prompt templates used by each module are reported in Appendix~\ref{app:prompts}.

\noindent\textbf{Online Semantic Event Construction.}
As described in Section~\ref{sec:method}, the input video is processed as a stream of non-overlapping 2-second chunks. Here, we provide additional details on the binary decision used during continuity checking. After a new active event is initialized, the current chunk is captioned using the prompt reported in~\ref{app:prompts_segmentation}; this caption serves as the anchor description of the event. The model then examines each subsequent chunk using the continuation prompt in~\ref{app:prompts_segmentation} and decides whether the chunk belongs to the same event or starts a new one.
Concretely, we score the two admissible responses using the model's next-token generation probabilities. Given the anchor caption $\gamma_{\mathrm{anc}}$, we define
\begin{align}
\epsilon_{\texttt{same}} &=
P_{\Phi_{\mathrm{LMM}}}\!\left(
\texttt{same} \mid S_m, \pi_{\mathrm{bin}}(\gamma_{\mathrm{anc}})
\right), \\
\epsilon_{\texttt{new}} &=
P_{\Phi_{\mathrm{LMM}}}\!\left(
\texttt{new} \mid S_m, \pi_{\mathrm{bin}}(\gamma_{\mathrm{anc}})
\right).
\end{align}
$S_m$ is assigned to the active event when
$\epsilon_{\texttt{same}} \geq \epsilon_{\texttt{new}}$,
and a new event boundary is introduced when
$\epsilon_{\texttt{new}} > \epsilon_{\texttt{same}}$.

Upon detecting a boundary, the active event is closed and emitted as a semantically coherent event segment $c_i$. The incoming segment $S_m$ then starts a new active event, serving as its anchor segment; a new anchor caption is generated from it, and the same online procedure is repeated.

In our implementation, the exact verbalizers are selected to ensure that both admissible responses correspond to single-token outputs for the LMM being used. For Qwen2.5-VL and Qwen3-VL, we use the uppercase verbalizers \texttt{SAME} and \texttt{NEW}. For InternVL2.5, we instead use lowercase \texttt{same} and \texttt{new}, since uppercase \texttt{SAME} is split into multiple tokens, while lowercase \texttt{same} and \texttt{new} are single-token outputs. This allows all continuity decisions in our experiments to be computed directly from the next-token distribution. A step-by-step pseudocode description of the Online Semantic Event Construction procedure is given in Algorithm~\ref{alg:event_segmentation}.

\begin{algorithm}[t]
\caption{Online Semantic Event Construction}
\label{alg:event_segmentation}
\begin{algorithmic}[1]
\Require Video stream $\mathcal{S}=\{S_t\}_{t=1}^{T}$, LMM $\Phi_{\mathrm{LMM}}$, captioning prompt $\pi_{\mathrm{cap}}$, binary prompt $\pi_{\mathrm{bin}}(\cdot)$
\Ensure Event sequence $\mathcal{C}_{1:m_T}=\{c_1,\dots,c_{m_T}\}$

\State $\mathcal{C} \gets \emptyset$, \quad $m_T \gets 0$
\Comment{Set of completed event segments}
\State $t \gets 1$

\While{$t \leq T$}
    \State $c \gets [S_t]$
    \Comment{Initialize a new active event}
    \State $\gamma_{\mathrm{anc}} \gets \Phi_{\mathrm{LMM}}(S_t,\pi_{\mathrm{cap}})$
    \Comment{Caption the anchor segment}
    \State $t \gets t + 1$

    \While{$t \leq T$}
        \State $\epsilon_{\texttt{same}} \gets
        P_{\Phi_{\mathrm{LMM}}}\!\left(
        \texttt{same} \mid S_t,\pi_{\mathrm{bin}}(\gamma_{\mathrm{anc}})
        \right)$
        \State $\epsilon_{\texttt{new}} \gets
        P_{\Phi_{\mathrm{LMM}}}\!\left(
        \texttt{new} \mid S_t,\pi_{\mathrm{bin}}(\gamma_{\mathrm{anc}})
        \right)$
        \If{$\epsilon_{\texttt{same}} \geq \epsilon_{\texttt{new}}$}
            \State $c \gets c \cup [S_t]$
            \Comment{Append the current segment to the active event}
            \State $t \gets t + 1$
        \Else
            \State \textbf{break}
            \Comment{A new event starts at $S_t$}
        \EndIf
    \EndWhile

    \State $m_T \gets m_T + 1$
    \State $c_{m_T} \gets c$
    \Comment{Store the completed event segment}
    \State $\mathcal{C} \gets \mathcal{C} \cup \{c_{m_T}\}$
\EndWhile

\State \Return $\mathcal{C}_{1:m_T}=\{c_1,\dots,c_{m_T}\}$
\end{algorithmic}
\end{algorithm}

\noindent\textbf{Zero-shot Event-to-Step grounding}
For each event \(E_j\), baseline step identification is performed as a multi-choice classification problem using the prompt in Appendix~\ref{app:prompts_classification}. The candidate set includes all task steps together with an additional "None of the above" option. The probability of the step is estimated by taking the probability of the model generating the letter token corresponding to the candidate option.

\noindent\textbf{Hierarchical Refinement and Substep Reweighting}
During refinement, the current event segment is evaluated against all candidate substeps using the prompt in Appendix~\ref{app:prompts_refinement}. Each candidate is formatted as \texttt{<parent\_step:substep>} so that the model can jointly consider the fine-grained action and its high-level procedural context. We also include the original parent-step description as an additional candidate, allowing the model to select the high-level step directly when the generated substeps are unnecessary or less informative. 

\section{Prompts}
\label{app:prompts}
This section reports the exact prompt templates used in the proposed pipeline. 

\subsection{Prompts for Online Event Segmentation}
\label{app:prompts_segmentation}
To construct semantically consistent event segments, each incoming 2-second video chunk is first captioned using a dataset-specific description prompt. For HT-Step, we use:
\begin{tcolorbox}[
    enhanced,
    title={HTStep description prompt},
    colback=gray!10,
    colframe=black!75,
    colbacktitle=black!75,
    coltitle=white,
    fonttitle=\bfseries,
    boxrule=1pt,
    arc=2mm
]
\small
Describe the action being performed by the person in 8--10 words in this 2 second video segment. \\

Describe only what is directly observable in the video.
\end{tcolorbox}

For Ego4D Goal-Step, we use:
\begin{tcolorbox}[
    enhanced,
    title={Ego4D GoalStep description prompt},
    colback=gray!10,
    colframe=black!75,
    colbacktitle=black!75,
    coltitle=white,
    fonttitle=\bfseries,
    boxrule=1pt,
    arc=2mm
]
\small
The video shows the first view of a person performing an action. \\

Describe the action being performed by the person in 8--10 words in this 2 second video segment. \\

Describe only what is directly observable in the video.
\end{tcolorbox}

For CrossTask, we use:
\begin{tcolorbox}[
    enhanced,
    title={CrossTask description prompt},
    colback=gray!10,
    colframe=black!75,
    colbacktitle=black!75,
    coltitle=white,
    fonttitle=\bfseries,
    boxrule=1pt,
    arc=2mm
]
\small
Describe in detail the action being performed by the person in this 2 second video segment. \\

Describe only the action performed by the person.
\end{tcolorbox}

Given the previous event description, action continuity is determined with the following prompt:
\begin{tcolorbox}[
    enhanced,
    title={Continuation / new-event prompt},
    colback=gray!10,
    colframe=black!75,
    colbacktitle=black!75,
    coltitle=white,
    fonttitle=\bfseries,
    boxrule=1pt,
    arc=2mm
]
\small
Given the previous event description: \{previous\_description\}, does the current video segment show exactly the SAME event continuing, or a NEW event starting? \\

Respond ONLY with ``SAME'' if it is the same event or ``NEW'' if it is a new event.
\end{tcolorbox}

\subsection{Prompts for Zero-shot Event-to-Step Grounding}
\label{app:prompts_classification}

For each predicted event segment, baseline step classification is performed using the following multi-choice prompt for all the datasets:
\begin{tcolorbox}[
    enhanced,
    title={Baseline step-classification prompt},
    colback=gray!10,
    colframe=black!75,
    colbacktitle=black!75,
    coltitle=white,
    fonttitle=\bfseries,
    boxrule=1pt,
    arc=2mm
]
\small
You are watching a video segment of someone attempting to \{goal\}.\\

What is the main action being performed in this exact moment? \\

Options:\\
\{choices\}\\

Answer with only the letter label of the correct option (e.g., A, B, ..., Z, AA, AB, etc.), with no extra text.
\end{tcolorbox}

\subsection{Prompts for Hierarchical Step Refinement}
\label{app:prompts_refinement}

When baseline uncertainty is high, the event segment is re-evaluated against fine-grained substep candidates using the following prompt for all the datasets:
\begin{tcolorbox}[
    enhanced,
    title={Hierarchical refinement prompt},
    colback=gray!10,
    colframe=black!75,
    colbacktitle=black!75,
    coltitle=white,
    fonttitle=\bfseries,
    boxrule=1pt,
    arc=2mm
]
\small
You are watching a video segment of someone attempting to \{goal\}.\\

What is the main action being performed in this exact moment?\\

Each option contains a context and an action. Choose the option considering both the context and the action.\\

Options:\\
\{choices\}\\

Answer with only the letter label of the correct option (e.g., A, B, ..., Z, AA, AB, etc.), with no extra text.
\end{tcolorbox}

\subsection{Prompts for Offline Substep Generation}
\label{app:prompts_substeps}

Substeps are generated offline from the textual procedure annotations using the following prompt for all the datasets:
\begin{tcolorbox}[
    enhanced,
    title={Substep-generation prompt},
    colback=gray!10,
    colframe=black!75,
    colbacktitle=black!75,
    coltitle=white,
    fonttitle=\bfseries,
    boxrule=1pt,
    arc=2mm
]
\small
Task: Given the target step, the recipe goal/activity, and the ordered list of other steps of the recipe:
\begin{itemize}
    \item Decompose the target step into atomic meaningful substeps that are necessary to complete the step.
\end{itemize}

Output requirements:
\begin{itemize}
    \item Return JSON only.
    \item Use this exact schema: \{"substeps": ["...", "..."]\}
    \item Generate between 1 and 5 items.
    \item Every substep must be complete in meaning and contain at least a verb and an object; if missing from the original step, infer it from the goal/activity/context.
    \item Do not repeat the original target step.
    \item Do not generate substeps for any step other than the target step.
    \item If the target step is already very simple and cannot be decomposed, return the empty substeps array.
\end{itemize}

Target step: \{target\_step\}

Goal of the recipe / activity: \{goal\}

Ordered steps from the recipe:
\{ordered\_steps\}
\end{tcolorbox}

\subsection{Qualitative results}
\begin{figure}[ht]
    \centering

    \includegraphics[width=\linewidth]{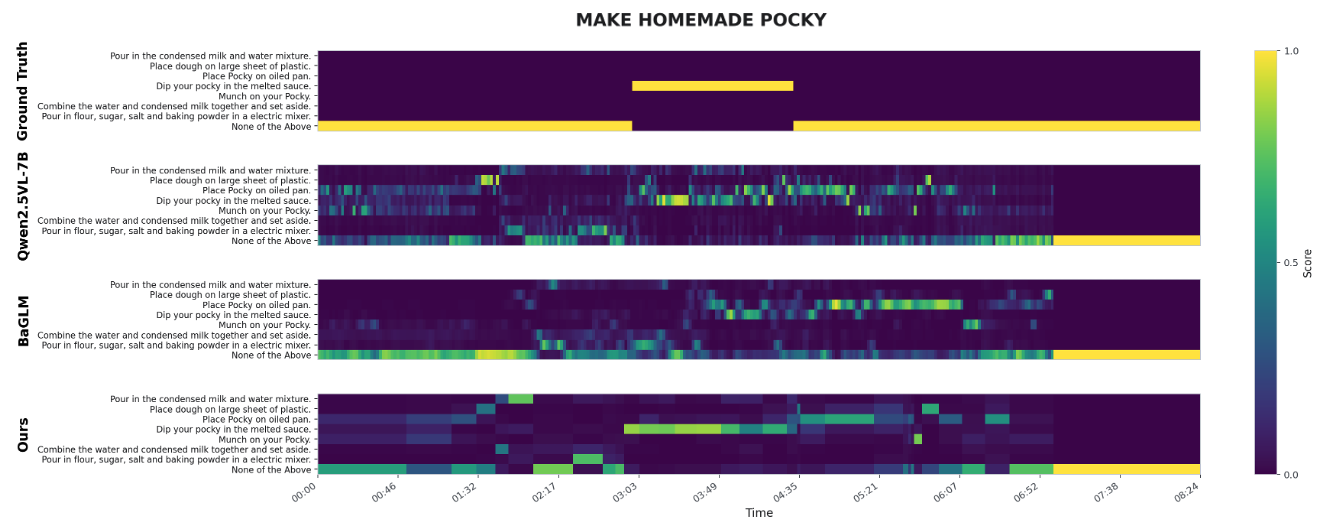}

    \vspace{0.5em}

    \includegraphics[width=\linewidth]{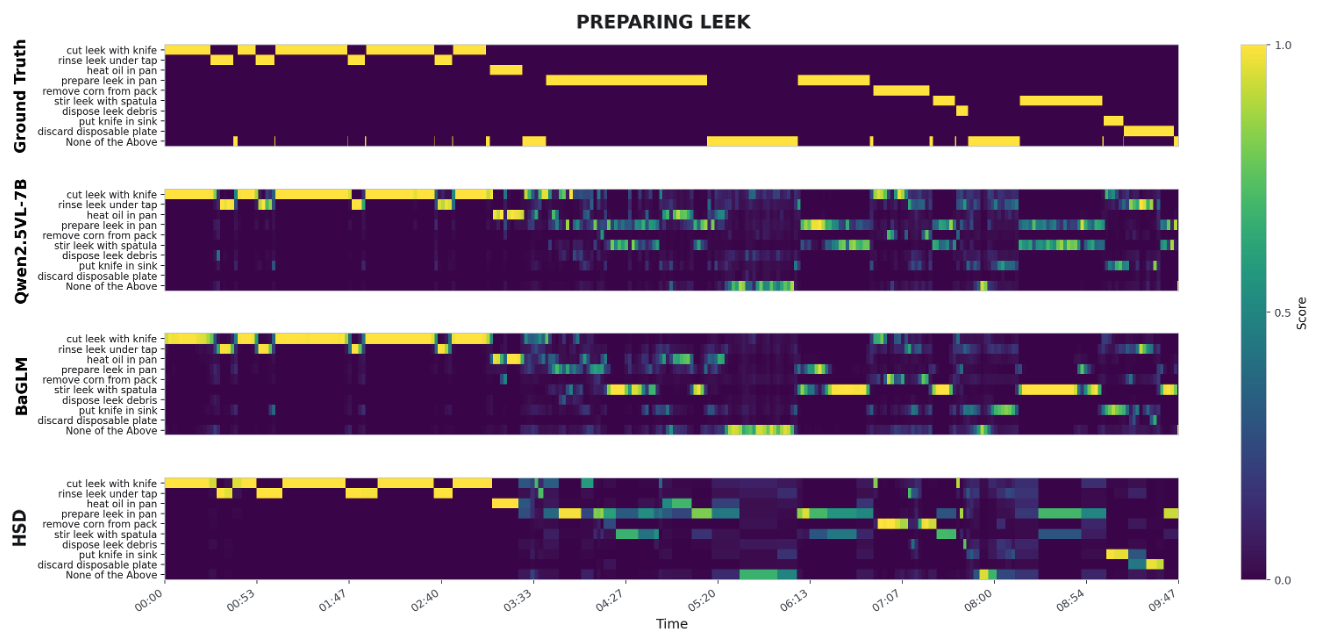}

    \caption{Qualitative results of \methodShort{} on videos from HT-Step (\emph{Make Homemade Pocky}) and Ego4D Goal-Step (\emph{Preparing leek}). Ground-truth step boundaries and predicted step probabilities per segment are shown for the baseline Qwen2.5-VL-7B, BAGLM with Qwen2.5-VL-7B, and \methodShort{} with the same backbone.}
    \label{fig:qualitative_htstep_ego4dgs}
\end{figure}

Figure~\ref{fig:qualitative_htstep_ego4dgs} presents additional full-video video qualitative comparisons one video from HT-Step and another one from Ego4D Goal-Step, showing ground-truth step boundaries together with the step probabilities predicted by the baseline Qwen2.5-VL-7B, BaGLM, and our method.

In the HT-Step example, \emph{Make Homemade Pocky}, the ground truth indicates that the relevant step, \emph{Dip your pocky in the melted sauce}, occurs only in the middle portion of the video. Qwen2.5-VL-7B produces noisy activations and assigns high probabilities to irrelevant steps around the true interval. BaGLM reduces some of this noise, but its predictions remain fragmented, particularly when distinguishing between visually or semantically similar steps such as \emph{Dip your pocky in the melted sauce} and \emph{Place pocky on oiled pan}. In contrast, our method concentrates probability on the correct step within its true temporal window.

In the Ego4D Goal-Step example, \emph{Preparing leek}, BaGLM also exhibits fragmented predictions. Although it captures part of the overall step ordering, its predictions frequently jump between neighboring steps and produce short, unstable activations that are not well aligned with the ground-truth boundaries. This is especially visible after the initial \emph{Cut leek with knife} / \emph{Rinse leek under tap} phase, where BaGLM switches among \emph{Heat oil in pan}, \emph{Prepare leek in pan}, and \emph{Stir leek with spatula}. Similar abrupt transitions appear later in the sequence, where BaGLM intermittently activates steps such as \emph{Remove corn from pack} and cleanup-related actions at incorrect timestamps. By contrast, our method produces more temporally stable predictions: it preserves the repeated early pattern, transitions more cleanly into the cooking phase, and avoids the back-and-forth fragmented outputs. As a result, its predictions are smoother and better aligned with the actual step boundaries throughout the video.

\begin{figure}[ht]
    \centering

    \includegraphics[width=\linewidth]{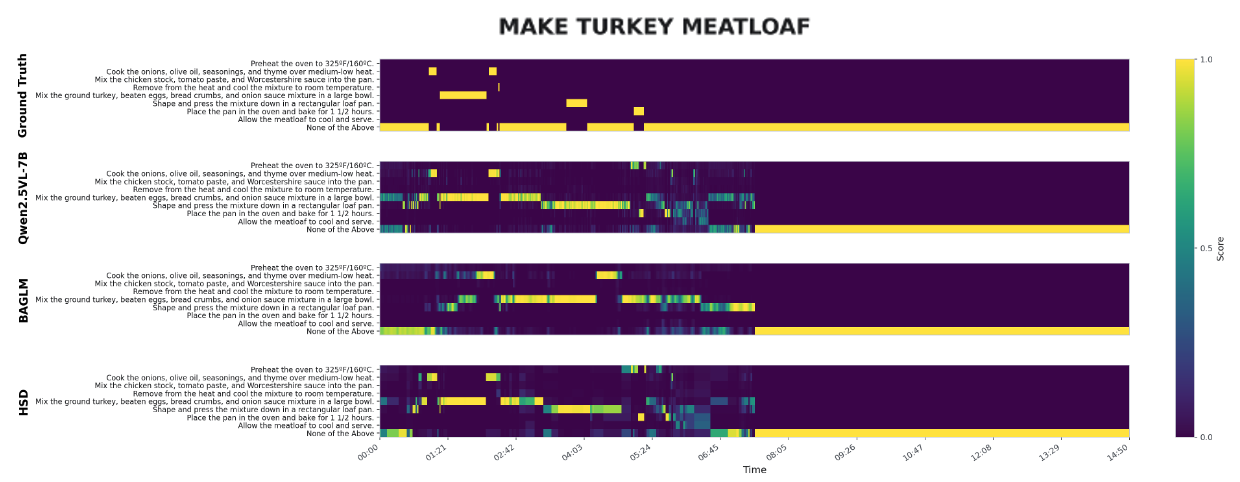}

    \vspace{0.5em}

    \includegraphics[width=\linewidth]{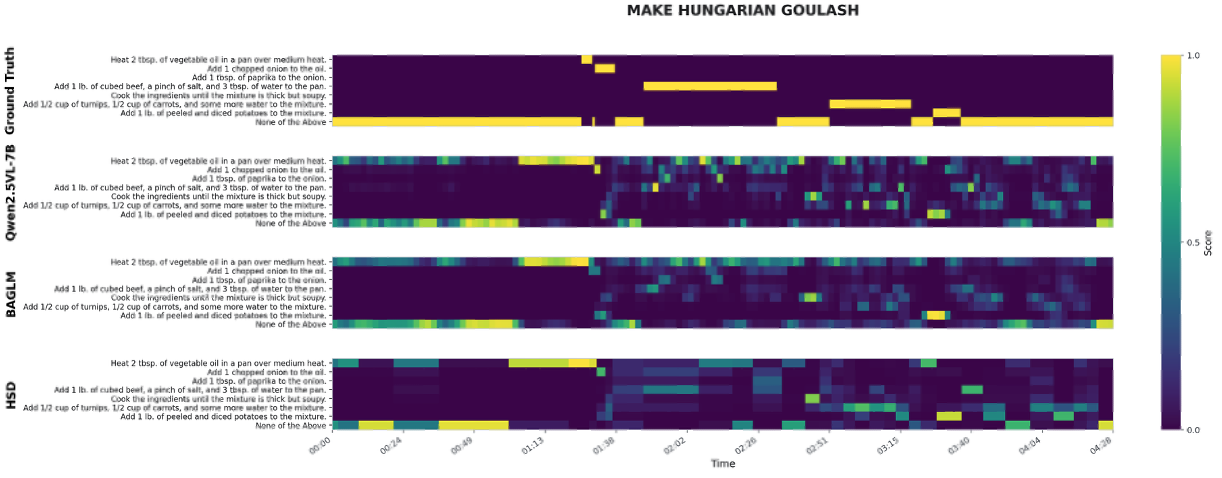}

    \caption{Qualitative results of \methodShort{} on two HT-Step videos: \emph{Make Turkey Meatloaf} and \emph{Make Hungarian Goulash}.}
    \label{fig:qualitative_htstep_extra}
\end{figure}

We further include two additional HT-Step examples, \emph{Make Turkey Meatloaf} and \emph{Make Hungarian Goulash}, in Figure~\ref{fig:qualitative_htstep_extra}. These examples show a similar overall trend: compared with BaGLM, our method produces more coherent temporal segments, reduces abrupt step switches, and better follows the ground-truth progression of the procedure. However, they also reveal a remaining limitation: our method does not always accurately capture the \emph{None of the above} option, occasionally assigning procedural steps to intervals that should be treated as background or non-step activity. This suggests that distinguishing valid procedural actions from irrelevant or transitional content remains an important direction for improvement.

\section{Broader Impact}
\label{App:broad_impact}
Our approach enables online understanding of instructional video steps without requiring task-specific training or additional annotated data. This makes it suitable for real-world scenarios where a system must process a video stream incrementally and provide timely feedback to users performing procedural tasks. Such capabilities can benefit applications in education, skill development, remote learning, cooking assistance, assembly guidance, and assistive technologies, especially in settings where annotated data is limited or immediate feedback is needed.

However, since our method builds on large pretrained multimodal models, it may inherit their limitations. These include biased or unreliable predictions, inconsistent performance across different activities, environments, cultures, or user populationss. In addition, online video understanding systems may raise privacy concerns if deployed in always-on or monitoring settings. Future work should investigate bias mitigation, privacy-preserving deployment and systematic evaluation across diverse users, activities, and environments.

\section{Assets}
Table~\ref{tab:assets} lists the main existing assets used in this work, including pretrained multimodal models and benchmark datasets. We use these assets in accordance with their corresponding licenses and terms of use. The datasets are used only for research evaluation, and the original creators are cited in the main paper.

\begin{table}[h]
\centering
\caption{List of URLs and licenses for the main assets used in this paper.}
\label{tab:assets}
\resizebox{\linewidth}{!}{
\begin{tabular}{lll}
\toprule
\textbf{Name} & \textbf{URL} & \textbf{License} \\
\midrule
Qwen2.5-VL-7B-Instruct &
\url{https://huggingface.co/Qwen/Qwen2.5-VL-7B-Instruct} &
Apache 2.0 \\

Qwen3-VL-8B-Instruct &
\url{https://huggingface.co/Qwen/Qwen3-VL-8B-Instruct} &
Apache 2.0 \\

InternVL2.5-8B &
\url{https://huggingface.co/OpenGVLab/InternVL2_5-8B} &
MIT \\

HT-Step &
\url{https://github.com/facebookresearch/htstep} &
CC-BY-NC 4.0 \\

Ego4D Goal-Step &
\url{https://github.com/facebookresearch/ego4d-goalstep} &
MIT \\

CrossTask &
\url{https://github.com/DmZhukov/CrossTask} &
BSD-3-Clause \\
\bottomrule
\end{tabular}
}
\end{table}

\section{Limitations} We identify two main limitations of our method. First, \methodShort{} relies on the visual-language reasoning capabilities of pretrained Large Multimodal Models for the three stages of the method. As a result, its effectiveness is closely tied to the quality of the underlying LMM. We expect that stronger LMMs will further improve the robustness and performance of our method. Second, our hierarchical refinement strategy depends on uncertainty estimation to decide when substep reasoning should be activated. While entropy-based activation is simple and effective, it is not perfect. Future work could explore more reliable refinement triggers, for example by incorporating temporal consistency.

\end{document}